\documentclass[smallcondensed,hidelinks]{svjour3}
\usepackage[utf8]{inputenc} 

\RequirePackage{fix-cm}
\smartqed  
\usepackage{graphicx}
\usepackage{mathptmx}      
\usepackage[linesnumbered, algo2e, vlined]{algorithm2e}
\usepackage{amsmath}
\usepackage{amsfonts}
\usepackage{amssymb}
\usepackage{url}
\usepackage{doi}
\renewcommand{\doitext}{\noexpand\textsc{DOI~}}
\usepackage{natbib}
\usepackage{booktabs}
\usepackage{hyperref}
\def\equationautorefname~#1\null{Equation~(#1)\null}%
\def\makeheadbox{\relax}

\begin{document}
\title{{\sc SLISEMAP}: Supervised dimensionality reduction through local explanations}
\titlerunning{{\sc SLISEMAP}}        
\author{
	Anton Bj\"orklund \and
	Jarmo M\"akel\"a \and
	Kai Puolam\"aki}
\authorrunning{A. Bj\"orklund et al.} 
\institute{A. Bj\"orklund \and J. M\"akel\"a \and K. Puolam\"aki \at
	University of Helsinki, Helsinki, Finland}
\date{}
\maketitle

\begin{abstract}
	Existing methods for explaining black box learning models often focus on building local explanations of model behaviour for a particular data item. It is possible to create global explanations for all data items, but these explanations generally have low fidelity for complex black box models.

	We propose a new supervised manifold visualisation method, {\sc slisemap}, that simultaneously finds local explanations for all data items and builds a (typically) two-dimensional global visualisation of the black box model such that data items with similar local explanations are projected nearby. We provide a mathematical derivation of our problem and an open source implementation implemented using the GPU-optimised PyTorch library.

	We compare {\sc slisemap} to multiple popular dimensionality reduction methods and find that {\sc slisemap} is able to utilise labelled data to create embeddings with consistent local white box models.
	We also compare {\sc slisemap} to other model-agnostic local explanation methods and show that {\sc slisemap} provides comparable explanations and that the visualisations can give a broader understanding of black box regression and classification models.

\end{abstract}

\section{Introduction}
\label{sec:intro}

In the past 20 years, manifold visualisation methods are a major development in the area of unsupervised learning. The trend that started from ISOMAP in 2000 \citep{isomap} has resulted in hundreds of methods to be developed, popular examples of which include methods such as t-SNE \citep{vandenmaaten2008} and UMAP \citep{2018arXivUMAP}. Manifold visualisation methods can be used to embed data into typically two or three dimensions while preserving some of the relevant features of the data. These methods have proven to be invaluable and central to exploring and understanding complex datasets in fields from genetics \citep{kobak2019,diaz2021}
to astronomy \citep{anders2018dissecting} and linguistics \citep{levine2020sensebert}.

Another recent development is explainable artificial intelligence (XAI), where the objective is
to understand and explore {\em{black box}} supervised learning algorithms; see \citet{guidotti:2018:a} for a recent survey. The explanation methods can roughly be divided into {\em global} and {\em local} methods.
Global methods try to explain the global behaviour of a supervised learning method by constructing a global understandable ({\em{white box}}) surrogate model that approximates the complex black box model. The drawback of the global approach is that for a sufficiently complex model, there is no simple surrogate model that would replicate the full model with a reasonable fidelity.

The alternative is local explanations that focus on how individual data items are classified or regressed. The advantage is that it is often possible to give high-fidelity interpretable local explanations, and the obvious disadvantage is that an explanation that is good for one data item may be useless for the other data items. A common model-agnostic approach for local explanations is to locally approximate the black box model with an interpretable white box model. These white box models are used to better understand the decision process by, e.g., showing which variables affect the outcome and how to achieve a different outcome.

In this paper, we combine the above two developments, namely, manifold visualisations and local explanations, to obtain {\em global} supervised manifold visualisations of the space of {\em local} explanations by using outputs of various black box supervised learning algorithms. We call the algorithm {\sc slisemap}.

The idea of {\sc slisemap} is straightforward: we want to find an embedding of data points into a (typically) two-dimensional plane such that the supervised learning model of the data points that are nearby in the embedding are explained by the same interpretable model.
The embedding of the data points and the local models associated with each point in the embedding form a global explanation of the supervised learning model as a combination of the local explanations. At the same time, our method produces a visualisation of the data where the data points that are being classified (or regressed) with the same rules are shown nearby.

\begin{figure}
	\centering
	\includegraphics[width=\textwidth]{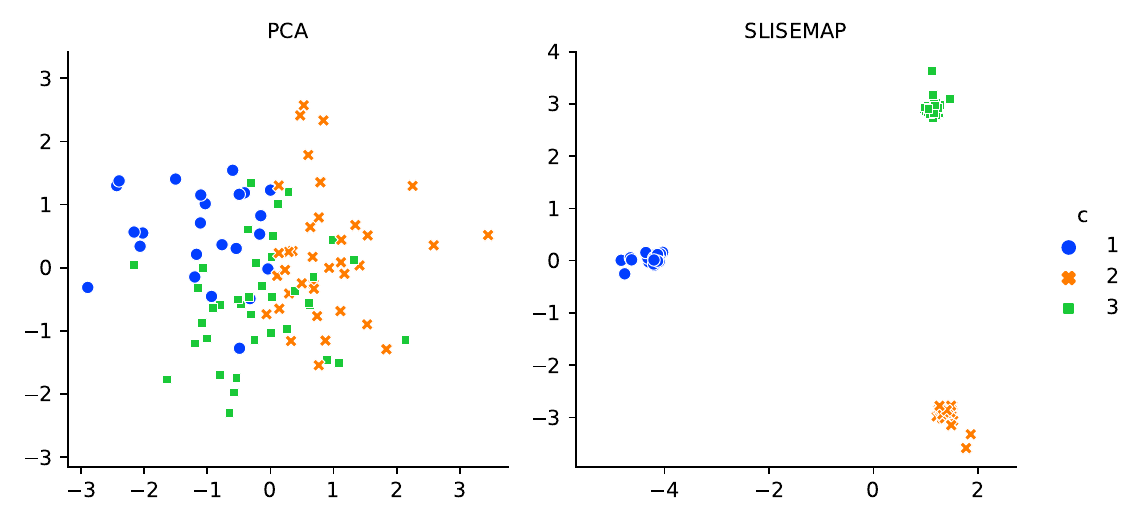}
	\caption{PCA (left) and {\sc slisemap} (right) embeddings of a toy dataset described in the text. The toy data matrix consists of $4$-dimensional Gaussian noise in ${\bf{X}}\in{\mathbb{R}}^{99\times 4}$, and the response vector ${\bf y} \in \mathbb{R}^{99}$ comes from a black box model $f({\bf{x}})=\max{\bf{x}}_{1:3}$.
	The legend in the plots corresponds to the value of ${\bf{c}}_i=\arg\max\nolimits_{j\in\{1,2,3\}}{{\bf{X}}_{ij}}$.
	We have added some jitter to the {\sc slisemap} embeddings to make the points in the clusters stand out.
	}
	\label{fig:toy}
\end{figure}

\paragraph{Example 1}
First, consider a toy regression example where we have $99$ data points composed of 4-dimensional covariates represented by rows of matrix ${\bf{X}}\in{\mathbb{R}}^{99\times 4}$ and a pretrained black box regression model given by function $f:{\mathbb{R}}^4\to{\mathbb{R}}$, which we want to study. The response vector ${\bf{y}}\in{\mathbb{R}}^{99}$ is given by the regression estimates as ${\bf{y}}_i=f({\bf{X}}_{i\cdot})$, where ${\bf{X}}_{i\cdot}$ denotes the $i$th row of the matrix ${\bf{X}}$. Unknown to the user, the elements of matrix ${\bf{X}}$ have been sampled at random from a normal distribution with zero mean and unit variance, and the regression function $f$ is given by $f({\bf{x}})=\max{{\bf x}_{1:3}}=\max\nolimits_{j\in\{1,2,3\}}{{\bf{x}}_j}$, where ${\bf{x}}\in{\mathbb{R}}^4$. In other words, the regression utilises the first three attributes in a nonlinear manner while ignoring the fourth attribute altogether.

Now, assume the user wishes to study the black box regression function and the dataset by embedding this 4-dimensional toy dataset into two dimensions.
Any dimensionality reduction method that only take the covariate matrix ${\bf{X}}$ into account, and ignore the response variables in ${\bf{y}}$, would see only Gaussian noise and result in a limited insight about the data and the regression function, as shown in the PCA visualisation of \autoref{fig:toy} (left).

Then, consider a variant of {\sc slisemap}, where ordinary least squares linear regression is used as an interpretable white box model. {\sc slisemap} will produce an embedding where the data are split into three clusters indexed by ${\bf{c}}_i=\arg{\max\nolimits_{j\in\{1,2,3\}}{{\bf{X}}_{ij}}}$, as shown in \autoref{fig:toy} (right). Each of the clusters corresponds to a different white box model denoted by $g_i:{\mathbb{R}}^4\to{\mathbb{R}}$ for all $i\in\{1,\ldots,99\}$ and are in this example simply given by $g_i({\bf{x}})={\bf{x}}_{{\bf{c}}_i}$.

For these toy data, {\sc slisemap} is therefore able to partition the data into three clusters, each modelled locally to a good accuracy by a separate linear white box model. The {\sc slisemap} embedding, together with the white box models, produces a global explanation of the black box model. The {\sc slisemap} embedding can be used to help the user to reverse-engineer the black box model and to find ``functional groups'' of data points, with each group modelled by a simple linear model.

Uninformative directions in the data space, such as the $4$th attribute in this example, are automatically ignored; {\sc slisemap} follows the possibly nonlinear manifold that is relevant for the supervised learning task. Note that in each of the three clusters in the {\sc slisemap} embedding, the value of the response variable ${\bf{y}}_i$ obeys an identical distribution: nearby points in the {\sc slisemap} embedding have similar white box models, not necessarily similar values of the response variables!

\paragraph{Example 2}
A property of local explanations is that there may be several explanations, with roughly equally good fidelity, for any given data point. Consider the toy dataset described above, but let us assume that the user wants to add a new point ${\bf x}'$ where some of the maximal variables are identical, ${\bf x}'_i={\bf x}'_j=\max{{\bf x}'_{1:3}}$, where $i,j\in\{1,2,3\}$ and $i\ne j$. This new point would fit equally well into both of the clusters $i$ and $j$ and, hence, has two potential local explanations. As shown later in the experiments, this also occurs with real datasets, and {\sc slisemap} can be used to reveal this ambiguity, unlike more traditional local explanation methods that output only one white box model.

\begin{figure}
	\centering
	\includegraphics[width=\textwidth]{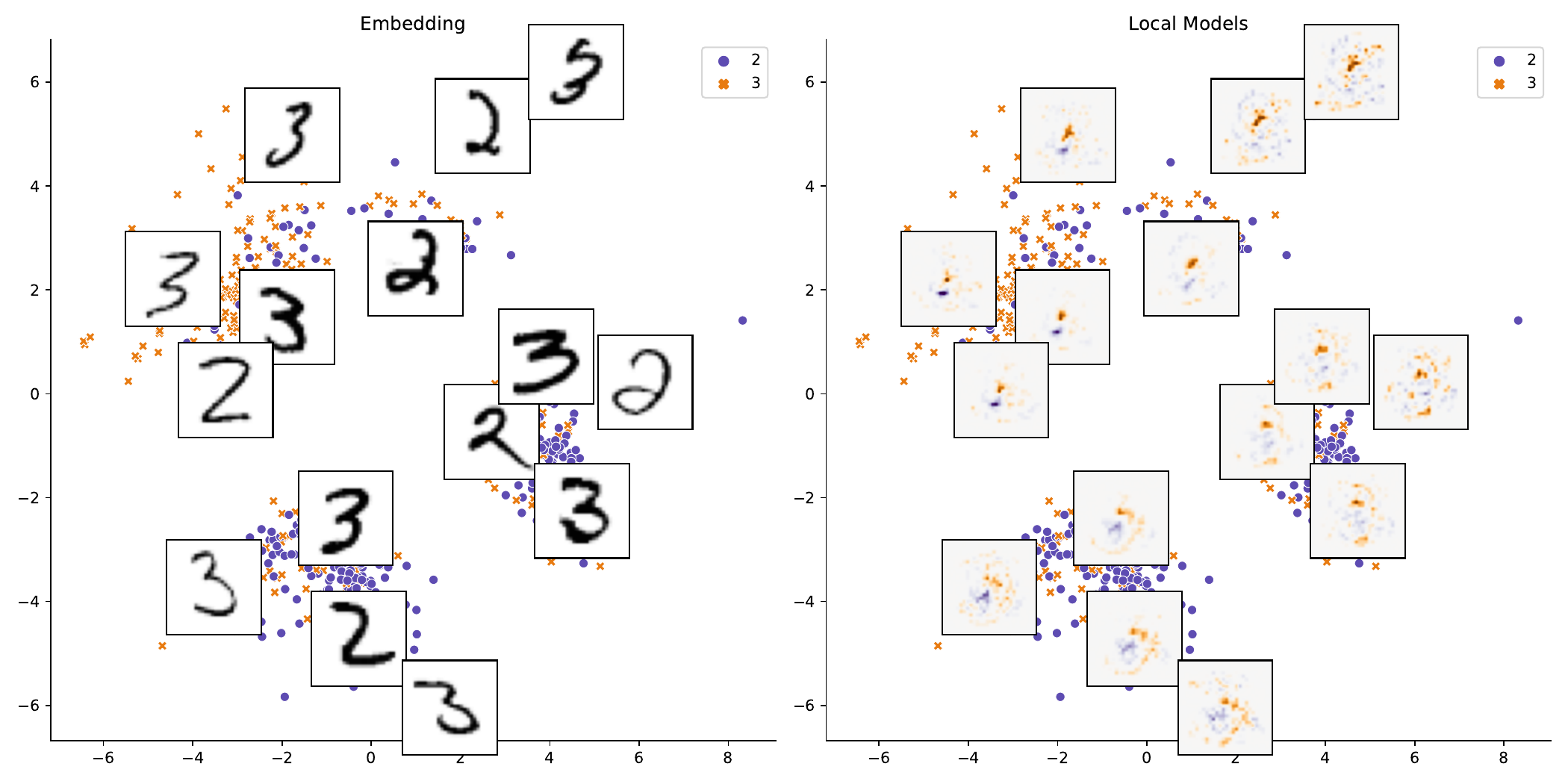}
	\caption{{\sc slisemap} visualisation of 2 s and 3 s in the MNIST dataset with a black box deep learning classifier that tries to classify the digits into 2 s and 3 s.
		The left shows the embedding of the digits in two dimensions, with a random selection of digits shown as images.
		The white box models are logistic regressions that use the image pixels as attributes.
		The right shows the same embedding, but the images show the regression coefficients associated with each pixel for the same selection of digits. The colour intensity indicates the magnitude of the coefficient. We can see from the right image that nearby digits are described by similar white box models.}
	\label{fig:mnist23}
\end{figure}

\paragraph{Example 3}
As a more realistic and complex example, Figure \ref{fig:mnist23} shows the visualisation of a black box model that classifies 2 versus 3 in the classic MNIST \citep{lecun1998gradientbased} dataset of hand-written digits. Here, the black box model is a convolutional neural network, and the white box model is a logistic regression classifier that takes the flattened image pixels as an input vector. The images are projected onto a two-dimensional plane such that the black box classifier for nearby images can, with good fidelity, be approximated by the same logistic regression model.
The digits are split into roughly four visually separable clusters, with digits in each of the clusters classified by different sets of pixels.
The logistic regression coefficients for different image pixels are shown in \autoref{fig:mnist23} (right). For example, the classifier separates the 2s and 3s at the bottom left mainly by identifying the "lower curve" in 3s, while in the top left, the classifier is looking for black pixels in the centre versus slightly below the centre. Seeing this visualisation enables us to find points that are similar in terms of the supervised learning problem and understand how the model classifies the data items.

The benefits of {\sc slisemap} compared to prior manifold visualisation or explanation methods include the following:
	(i) {\sc slisemap} finds visual patterns, like {\em clusters}, such that all data items within the same cluster are explained by the same simple model. For example, in Figure \ref{fig:toy} {\sc slisemap} reveals three clusters, while Figure \ref{fig:mnist23} shows roughly four clusters of digits that can be separated by a given subset of pixels.
	(ii) Unlike existing local explanation methods, {\sc slisemap} provides both {\em global} and {\em local explanations} of the data. For example, Figure \ref{fig:mnist23} compactly shows the explanations for all digits, in addition to the fact that roughly four linear models are sufficient to explain the classification of all digits to a reasonable fidelity.
	(iii) {\sc slisemap} can be used to discover a nonlinear structure in a dataset, as shown in Figure \ref{fig:mnist23} and later in \autoref{sec:vis}.

\subsection{Contributions}

The contributions of this paper are as follows:
(i) We define a criterion for a supervised manifold embedding that shows local explanations and give an efficient algorithm to find such embeddings.
(ii) We show experimentally that our method results in informative and useful visualisations and local white box models can be used to explain and understand supervised learning models.
(iii) We compare our contribution to manifold visualisation methods and comparable local explanation methods.

\section{Related work}
\label{sec:related}

In this section, we briefly review the explainable AI and dimensionality reduction methods.

\subsection{Explainable AI}
\label{sec:xai}

The explanations of black box models can be generally divided into the exploration of global aspects, i.e., the entire model \citep{baehrens:2010:a, henelius2014peek, henelius2017interpreting, adler:2016:a, datta:2016:a},
or inspection of local attributes, i.e., individual decisions \citep{ribeiro2016, ribeiro2018Anchors, fong:2017:a, Lundberg_Lee_2017};
See \citet{guidotti:2018:a} for a recent survey and references. On a global level, the scope of the explanations is on understanding {\em how} the model has produced predictions, where the {\em why} is usually beyond human comprehension due to model complexity.
On this level, we can examine which features affect the predictions most \citep{fisher2019AllMA} and what interactions there are between features \citep{Goldstein2013PeekingIT,henelius2014peek,henelius2017interpreting}.

However,
we are interested in local explanation methods, specifically those that can be used for any type of model (model-agnostic) and do not require any model modifications (post hoc). A common approach in this niche is to locally approximate the black box model with a simpler white box model.
One of the first such methods, {\sc lime} \citep{ribeiro2016}, generates interpretations for user-defined areas of interest by perturbing the data and training a linear model based on the predictions. Another similar method is {\sc shap} \citep{Lundberg_Lee_2017}, which finds the weights based on Shapley value estimation \citep{shapley1951}. Nonlinear white box models can also be used, such as decision rules \citep{guidotti2018Local,ribeiro2018Anchors}.

Many of these methods generate local explanations based on perturbed data, but designing a good data generation process is nontrivial \citep{guidotti:2018:a,laugel:2018, molnar:2019:a}, e.g., replacing pixels in an image with random noise seldom results in natural-looking images.
One method that only utilises existing data is called {\sc slise} \citep{bjorklund2019,bjorklundRobustRegressionError2022}, which finds the largest subset of data items that can be approximated (up to a given accuracy) by a sparse linear model.
The work presented here can be seen as a global extension of {\sc slise}.

\subsection{Dimensionality reduction}
\label{sec:dr}

Another way of assessing high-dimensional data is to reduce the number of covariates by, e.g., removing noninformative and redundant features or combining multiple features into single elements and thus making the data more interpretable. There are advantages of utilising dimensional reduction, as it removes correlated features in the data and allows for easier visualisation, e.g., in two dimensions, but combined features can also become less interpretable, and some information will inevitably be lost. The simplest dimensional reduction techniques are unsupervised methods operating on the whole dataset by keeping the most dominant features with, e.g., backward elimination and forward selection, or by finding a combination of new features.

These methods include principal component analysis (PCA) and other linear methods \citep{cunningham2015}.
Other approaches include 
locally linear embedding 
(LLE, MLLE) \citep{roweis2000nonlinear, zhang2007mlle}, spectral embedding \citep{belkin2003laplacian} and multidimensional scaling (MDS) \citep{kruskal1964multidimensional}, global-distance preserving MDS \citep{mead1992}, ISOMAP \citep{isomap}, t-SNE \citep{vandenmaaten2008}, and UMAP \citep{2018arXivUMAP}.
Recently, some supervised methods have also become available, based on t-SNE \citep{kang2020,hajderanj2019} and UMAP \citep{mcinnes2018umap-software}.

There are some recent developments towards combining dimensionality reduction with explainable AI.
\citet{anbtawi3DPlaygroundTSNE2019} presents an interactive tool, which embeds data with standard t-SNE and the user is able to examine the explanations of individual data-items created by {\sc lime}. However, there are no interactions between t-SNE and {\sc lime}. Meanwhile, \citet{DBLP:conf/esann/BibalVNF20} use {\sc lime} to explain the t-SNE embedding, with no supervised learning method involved.

\subsection{Local linear models}

Local linear models, such as \citet{nellesLocalLinearModel2000}, estimate a response variable by fitting linear models to neighbourhoods of data items. 
\citet{chengLocalLinearRegression2013} improves the computational efficiency by using dimensionality reduction, after which they apply local linear models on the embedding.
These methods use local models, similar to {\sc slisemap}. However, they are regression methods and do not produce visualisations nor explanations.

\section{Definitions and algorithms}
\label{sec:theory}

\subsection{Problem definition}
\label{sec:definitions}

A dataset consists of $n$ data points $({\bf x}_1,{\bf y}_1),\ldots,({\bf x}_n,{\bf y}_n)$, where ${\bf x}_i\in{\cal X}$ are the {\em covariates} and ${\bf y}_i\in{\cal Y}$ are \emph{responses} for one data point and $i\in[n]=\{1,\ldots,n\}$.
${\cal X}$ and ${\cal Y}$ are the domains of the covariates and responses, respectively. In this paper and in our software implementation, we restrict ourselves to real spaces, ${\cal X}={\mathbb{R}}^m$ and ${\cal Y}={\mathbb{R}}^p$, but the derivations in this subsection are general and would be valid, for example, for categorical variables as well.

The goal is to find a local {\em white box} model $g_i:{\cal X}\to{\cal Y}$ for every data point $({\bf x}_i, {\bf y}_i)$, where $i\in[n]$.
We use $\tilde{\bf y}_{ij}=g_i({\bf x}_j)$ to denote the estimate of ${\bf y}_j$ obtained by a white box model associated with data point ${\bf x}_i$.
Again, while the derivation is general, in this paper, we focus on cases where the white box model, $g_i$, is either a linear projection (for regression problems) or multinomial logistic regression (for classification problems), as defined later in Section \ref{sec:regclass}.

If we have access to a trained \emph{black box} supervised learning algorithm $f:{\cal X}\to{\cal Y}$, then we can use estimates given by the model $\hat{\bf y}_i=f({\bf x}_i)$ instead of ${\bf y}_i$. This will make the local models $g_i$ local approximations of the black box model. These approximations can then also be used to explain the predictions of the black box model as in \cite{bjorklund2019}.

Additionally, we want to find a lower-dimensional embedding ${\bf Z}_{i\cdot}$, where ${\bf Z}\in\mathbb{R}^{n \times d}$ for every data point $i\in[n]$ and where ${\bf Z}_{i\cdot}$ denotes the $i$th row of matrix ${\bf Z}$. Our objective is that neighbouring data items in the embedding space have similar local models $g_i$. Since we focus on visualisation, in our examples, ${\bf Z}_{i\cdot}$ is typically 2-dimensional ($d=2$).

We denote by ${\bf D}_{ij}$ the Euclidean distance between the points ${\bf Z}_{i\cdot}$ and ${\bf Z}_{j\cdot}$ in the embedding, where
\begin{equation}
	\label{eq:D}
	{\bf D}_{ij}=\left(\sum\nolimits_{k=1}^d{\left({\bf Z}_{ik}-{\bf Z}_{jk}\right)^2}\right)^{1/2}.
\end{equation}
We define the {\em soft neighbourhood} by using a softmax function as follows:
\begin{equation}
	\label{eq:W}
	{\bf W}_{ij}=\frac{e^{-{\bf D}_{ij}}}{\sum\nolimits_{k=1}^n{e^{-{\bf D}_{ik}}}}.
\end{equation}
We define the {\em{radius}} of the $d$-dimensional embedding to be the square root of the variance of the embedding or
\begin{equation}\label{eq:radius}
	{\textrm{radius}}({\bf Z})=\left(
	\frac{1}{n}\sum_{i=1}^{n}{\sum_{k=1}^{d}{{\bf Z}_{ik}^2}}
	\right)^{1/2}.
\end{equation}

We further define a loss function $l:{\cal Y}\times{\cal Y}\to{\mathbb{R}_{\ge0}}$ for the white box models. Here, we use the shorthand notation
\begin{equation}\label{eq:Lij}
	{\bf L}_{ij}=l(\tilde{\bf y}_{ij},{\bf y}_j)=l(g_i({\bf x}_j),{\bf y}_j).
\end{equation}
In this work, we use quadratic losses (for regression) and Hellinger distances between multinomial distributions (for classification), which we define later in Section \ref{sec:regclass}.

The local white box model $g_i$ can optionally have a regularisation term, which we denote by $G_i$.
Since {\sc slisemap} consists of local models, regularisation can be important to handle small neighbourhoods.
In this paper, we will use Lasso regularisation \citep{roberttibshiraniRegressionShrinkageSelection1996} to be later defined in Equations \eqref{eq:lassor} and \eqref{eq:lassoc} in Section \ref{sec:regclass}.

Recall that the goal is that all points in the (soft) neighbourhood of point ${\bf Z}_{i\cdot}$ to be modelled well by the local white box model $g_i$. Mathematically, this can be formalised as minimising the following weighted loss:
\begin{equation}\label{eq:Li}
	{\cal L}_i=\sum\nolimits_{j=1}^n{{\bf W}_{ij}{\bf L}_{ij}}+G_i,
\end{equation}
Each local model $g_i$ has its own set of weights ${\bf W}_{i\cdot}$, of which ${\bf W}_{ii}$ is the largest (due to ${\bf D}_{ii} = 0$).
This is what makes the models \emph{local}. If the embedding, and therefore ${\bf W}$, is fixed, we can obtain the local models simply by minimising the loss of Equation \eqref{eq:Li}.

Our final loss function is obtained by summing all losses given by Equation \eqref{eq:Li}.
We summarise everything in the main problem definition:
\begin{problem}\label{prob:slisemap}
{\sc slisemap}
Given dataset $({\bf x}_1,{\bf y}_1),\ldots,({\bf x}_n,{\bf y}_n)$, white box functions $g_i$ and regularisation terms $G_i$ for $i\in[n]$, loss function $l$, and the desired radius of the embedding $z_{radius}>0$, find the parameters for $g_1,\ldots,g_n$ and embedding ${\bf Z}\in{\mathbb{R}}^{n\times d}$ that minimise the loss given by
\begin{equation}\label{eq:main}
	{\cal L}=
	\sum_{i=1}^n{\sum_{j=1}^n{{\bf W}_{ij}{\bf L}_{ij}}}
	+\sum_{i=1}^n{G_i}.
\end{equation}
where ${\bf L}_{ij}=l(g_i({\bf x}_j),{\bf y}_j)$,
	${\bf W}_{ij}=e^{-{\bf D}_{ij}}/\sum\nolimits_{k=1}^n{e^{-{\bf D}_{ik}}}$, and ${\bf D}_{ij}=(\sum\nolimits_{k=1}^d{({\bf Z}_{ik}-{\bf Z}_{jk})^2})^{1/2}$, with the constraint that ${\rm{radius}}({\bf Z}) = z_{radius}$.
\end{problem}

The loss function is invariant with respect to the rotation, which means that the embedding is invariant under rotation.
The $z_{\rm radius}$ parameter essentially fixes the sizes of the neighbourhoods. At the limit of small $z_{\rm radius}$, all points will be compressed close to the origin, and hence, all points will be described by the same local model. On the other hand, if $z_{\rm radius}$ is very large, the points are far away from each other, and the neighbourhood of each of the points consists only of the point itself.

\subsection{Adding new data points to an existing solution}
\label{sec:adding}

Often, it is useful to add new data points to an existing embedding without recomputing the whole embedding. Here, we define an auxiliary problem to this end.

Assume that we have a new data point denoted by $({\bf x}_{n+1},{\bf y}_{n+1})$. Define parameters for a new local model $g_{n+1}$ and a new embedding matrix by ${\bf Z}'\in{\mathbb{R}}^{(n+1)\times d}$, such that the first $n$ rows are the solution to \autoref{prob:slisemap}.
We formulate the problem of adding a new point to an existing {\sc slisemap} solution as follows:

\begin{problem}\label{prob:add}
{\sc slisemap-new}
Given the definitions above and a new data point $({\bf x}_{n+1},{\bf y}_{n+1})$, find the parameters for $g_{n+1}$ and ${\bf Z}_{n+1,\cdot}'\in{\mathbb{R}}^d$ such that the loss of Equation~\eqref{eq:main} is minimised; when is added to the set of local models and ${\bf Z}$ is replaced by ${\bf Z}'$.
\end{problem}

Solving Problem \ref{prob:add} is much easier than solving the full Problem \ref{prob:slisemap} because in Problem \ref{prob:add}, only the parameters for the new point need to be found, as opposed to the parameters for the $n$ points in the full \autoref{prob:slisemap}. As a drawback, solving the full problem should result in slightly smaller loss. However, the difference should asymptotically vanish at the limit of large $n$. We study this difference experimentally in \autoref{sec:subset}.

\subsection{Slisemap for regression and classification}
\label{sec:regclass}

While the definitions in Section \ref{sec:definitions} were general, in this paper, we focus on regression and classification problems where the covariates are given by $m$-dimensional real vectors, or ${\cal X}={\mathbb{R}}^m$. We denote the data matrix by ${\bf X}\in{\mathbb{R}}^{n\times m}$, where the rows correspond to the covariates or ${\bf X}_{i\cdot}={\bf x}_i$. If necessary, we include in the data matrix a column of ones to account for the intercept terms.

\paragraph{Regression} In regression problems, we use linear regression as the white box model. More specifically, we assume that the dependent variables are real numbers or ${\cal Y}={\mathbb{R}}$. The white box regression model is given by a linear function
\begin{equation}\label{eq:gR}
	g_R({\bf x},{\bf b})={\bf x}^T{\bf b},
\end{equation}
where ${\bf b}\in{\mathbb{R}}^m$, and the loss is quadratic,
\begin{equation}\label{eq:LR}
	l_R(\tilde{\bf y},{\bf y})=\left(\tilde{\bf y}-{\bf y}\right)^2.
\end{equation}

The linear regression model $g_R$ is parametrised by the vector ${\bf b} \in \mathbb{R}^m$. If we gather the parameter vectors from all the local models in \autoref{prob:slisemap} into one matrix ${\bf B} \in \mathbb{R}^{n \times m}$ such that the row ${\bf B}_{i\cdot}$ gives the parameter vector of the local model $g_i$, then the parameters being optimised in \autoref{prob:slisemap} are ${\bf B}$ and ${\bf Z}$.

We use Lasso regularisation, see \autoref{eq:Li}, for any $i\in[n]$ given by
\begin{equation}\label{eq:lassor}
	G_i^R=\lambda\times\sum_{j=1}^m{\left|{\bf B}_{ij}\right|},
\end{equation}
where $\lambda$ is a parameter setting the strength of the regularisation.
We can then write Equation \eqref{eq:main} to be optimised explicitly as
${\cal L}_R({\bf X},{\bf y},{\bf B},{\bf Z})$
with ${\bf L}_{ij}=\left(({\bf X}{\bf B}^T)_{ij}-{\bf y}_j\right)^2$.

\paragraph{Classification} In classification problems, we assume that the black box classifier outputs class probabilities for $p$ classes. We use multinomial logistic regression as the white box model. The dependent variables are multinomial probabilities in $p$-dimensional simplex or ${\cal Y}=\{{\bf y}\in{\mathbb{R}}^p_{\ge 0}\mid\sum\nolimits_{i=1}^p{{\bf y}_i}=1\}$. Multinomial logistic regression can be parametrised by ${\bf b} \in \mathbb{R}^{(p-1) m}$. The white box classification model is that of the multinomial logistic regression \citep{hastie2009},
\begin{equation}\label{eq:gC}
	\tilde{\bf y}_i=
	g_C({\bf x},{\bf b})_i=
	\begin{cases}
		\frac{\exp\left({{\bf x}^T{\bf b}_{((i-1)m+1):(im)}}\right)}
		{1+\sum\nolimits_{j=1}^{p-1}{\exp\left({{\bf x}^T{\bf b}_{((j-1)m+1):(jm)}}\right)}}
		 & \text{if $i<p$} \\
		\frac{1}
		{1+\sum\nolimits_{j=1}^{p-1}{\exp\left({{\bf x}^T{\bf b}_{((j-1)m+1):(jm)}}\right)}}
		 & \text{if $i=p$}
	\end{cases},
\end{equation}
We used ${\bf b}_{a:b}$ to denote an $(b-a+1)$-dimensional vector $({\bf b}_a,{\bf b}_{a+1},\ldots,{\bf b}_b)^T$.
When using $g_C$ as the white box model in \autoref{prob:slisemap}, we can express the parameters for all the local models using a matrix ${\bf B} \in \mathbb{R}^{n \times (p-1)m}$, where the $i$th row ${\bf B}_{i\cdot}$ corresponds to the parameter vector of the $i$th data point.

The loss function could be any distance measure between multinomial probabilities, such as Kullback-Leibler (KL) divergence. Here, however, we choose the more numerically stable squared Hellinger distance \citep{ali1966,liese2006},
\begin{equation}\label{eq:LC}
	l_C(\tilde{\bf y},{\bf y}) =
	\frac 12\sum_{i=1}^p{\left(\sqrt{\tilde{\bf y}_i}-\sqrt{{\bf y}_i}\right)^2} =
	1-\sum_{i=1}^p{\sqrt{\tilde{\bf y}_i{\bf y}_i}}.
\end{equation}
The squared Hellinger distance is symmetric and bounded in interval $[0,1]$, unlike the KL, which is not symmetric or upper bounded. The squared Hellinger distance has convenient information-theoretic properties; for example, it is proportional to a tight lower bound for the KL divergence.

Note that when there are only two classes ($p=2$), the multinomial logistic regression reduces to the standard logistic regression.

As in the regression formulation, we use Lasso regularisation for $i\in[n]$ given by
\begin{equation}\label{eq:lassoc}
	G_i^C=\lambda\times\sum_{j=1}^{(p-1)m}{\left|{\bf B}_{ij}\right|}.
\end{equation}
where $\lambda$ is a parameter setting the strength of the regularisation.

We can then write Equation \eqref{eq:main} to be explicitly optimised as
${\cal L}_C({\bf X},{\bf y},{\bf B},{\bf Z})$
with ${\bf L}_{ij}=l_C(g_{Ci}({\bf x}_j),{\bf y}_j)$ expressed by using the Hellinger loss $l_C$ of Equation \eqref{eq:LC} and multinomial logistic regression $g_C$ of Equation \eqref{eq:gC}.

\paragraph{Alternative formulation for binary classification} In case the targets are given by a black box model, we can also use an alternative formulation for binary classification ($p=2$).
Here, we simply transform the probability $\hat y_1$ with a logit function, $\hat y_1' = \log(\hat y_1 / (1 - \hat y_1))$, from the interval $[0,1]$ to the interval $[-\infty,\infty]$ and then run {\sc slisemap} for regression with quadratic loss, as above.
Using a logit transformation followed by a linear model matches the behaviour of {\sc shap} \citep{Lundberg_Lee_2017} and {\sc slise} \citep{bjorklund2019}.

\subsection{Algorithm}
\label{sec:algo}

Pseudocode for {\sc slisemap} is given in \autoref{alg:slisemap}.
As the initial values for the embedding ${\bf Z}$, we use the principal component projection of the data (PCA).
Then, we optimise the values of ${\bf B}$ and ${\bf Z}$ by minimising the loss given by \autoref{eq:main}.

\begin{algorithm2e}[]
	\DontPrintSemicolon
	\SetKwProg{Fn}{Function}{}{}
	\SetKwFunction{slisemapf}{Slisemap}
	\SetKwFunction{slisemapnewf}{Slisemap-new}
	\SetKwFunction{escapef}{Escape}
	\SetKwFunction{softmaxf}{Softmax}
	\SetKwFunction{radiusf}{Radius}
	\SetKwRepeat{Do}{do}{while}

	\Fn{
	\slisemapf{${\bf X}$, ${\bf y}$, $z_{\rm radius}$, $d$}
	}{
	\tcc{Use the $d$ first principal components as the initial embedding}
	${\bf Z} \gets \textrm{PCA}({\bf X})_{\cdot,1:d}$\;
	${\bf Z}\gets {\bf Z}/{\rm{radius}}({\bf Z})$\tcc*{Normalise ${\bf Z}$ using \autoref{eq:radius}}
	\tcc{Optimise ${\bf B}$ (only) as the initial parameter matrix}
	${\bf B} \gets {\arg\min}_{{\bf B}}{\left[{\cal L}({\bf X}, {\bf y}, {\bf B}, {\bf Z}\times z_{\rm radius})\right]}$\;
	\Do{not converged}{
	\tcc{Shuffle points to escape local minima}
	${\bf B}_{i\cdot}', {\bf Z}_{i\cdot}' \gets \escapef({\bf X}_{i\cdot},{\bf y}_i, {\bf B}, {\bf Z}\times z_{\rm radius}/{\textrm{radius}}({\bf Z}))$ for all $i\in[n]$\;
	${\bf B},{\bf Z}\gets{\bf B}',{\bf Z}'$\;
	\tcc{Use L-BFGS to optimise \autoref{eq:main}}
	${\bf B}, {\bf Z} \gets {\arg\min}_{{\bf B},{\bf Z}}{\left[
		{\cal L}({\bf X}, {\bf y}, {\bf B}, {\bf Z}\times z_{\rm radius}/{\textrm{radius}}({\bf Z}))
		+({\textrm{radius}}({\bf Z})-1)^2
		\right]}$\;
	}
	${\bf Z}\gets {\bf Z}\times z_{\rm radius}/{\textrm{radius}}({\bf Z})$ \tcc*{Set the radius of ${\bf Z}$ using \autoref{eq:radius}}
	\KwResult{${\bf B}$, ${\bf Z}$}
	}

	\vspace*{0.5em}

	\Fn{
	\escapef{${\bf x}'$,${\bf y}'$, ${\bf B}$, ${\bf Z}$}
	}{
	Compute the weight matrix ${\bf W}$ from ${\bf Z}$ by using Equations \eqref{eq:D} and \eqref{eq:W}\;
	${\bf l}_i\gets l(g_i({\bf x}'),{\bf y}')$ for all $i\in[n]$\tcc*{\autoref{eq:Lij}}
	$k \gets {\arg\min}_k \sum\nolimits_{j=1}^n {\bf W}_{kj} {\bf l}_j$\;
	\KwResult{${\bf B}_{k\cdot}, {\bf Z}_{k\cdot}$}
	}

	\vspace*{0.5em}

	\Fn{
	\slisemapnewf{${\bf x}_{new}$, ${\bf y}_{new}$, ${\bf X}_{old}$, ${\bf y}_{old}$, ${\bf B}_{old}$, ${\bf Z}_{old}$, $z_{\rm radius}$}
	}{
	${\bf b}', {\bf z}' \gets \escapef({\bf x}_{new}, {\bf y}_{new}, {\bf B}_{old}, {\bf Z}_{old})$\;
	${\bf b}_{new}, {\bf z}_{new} \gets {\arg\min}_{{\bf b}', {\bf z}'}{ {\cal L}(
		\left[{{\bf X}_{old} \atop {\bf x}_{new}}\right],
		\left[{{\bf y}_{old} \atop {\bf y}_{new}}\right],
		\left[{{\bf B}_{old} \atop {\bf b}'}\right],
		\left[{{\bf Z}_{old} \atop {\bf z}'}\right] \times z_{\rm radius} / {\rm radius}(\left[{{\bf Z}_{old} \atop {\bf z}'}\right])
		)
		}$\;
	\KwResult{${\bf b}_{new}$, ${\bf z}_{new}$}
	}

	\caption{The {\sc slisemap} algorithm, where ${\cal L}$ is given in \autoref{eq:main}, ${\bf W}$ in \autoref{eq:W} and ${\rm{radius}}({\bf Z})$ in \autoref{eq:radius}. See the text for discussion.}
	\label{alg:slisemap}
\end{algorithm2e}

In our algorithm, we keep the radius of the embedding ${\bf Z}$ constant by always dividing it by ${radius}({\bf Z})$ during the optimisation. Due to this normalisation, the loss term ${\cal L}()$ does not depend on the radius of ${\bf Z}$. Thus, for \emph{numerical stability}, we add a small penalty term $({radius}({\bf Z})-1)^2$ to the loss (line 8 of Algorithm \ref{alg:slisemap}).

For the implementation of ``$\arg\min$'' in Algorithm \ref{alg:slisemap}, we use PyTorch \citep{pytorch}, which enables us to \emph{optionally} take advantage of GPU acceleration. The optimisation of ${\bf B}$ and ${\bf Z}$ is performed using the L-BFGS \citep{nocedal1980updating} optimiser of PyTorch.
As explained earlier, in this paper, we assume that the data are real valued and use the white box models and losses of Section \ref{sec:regclass} to study regression and classification problems.

In addition to the L-BFGS gradient search, we use an additional heuristic (function \texttt{Escape} in Algorithm \ref{alg:slisemap}) to help with escaping local optima. The heuristic consists of moving each item (embedding and local model) to the soft neighbourhood, given by ${\bf W}$ in \autoref{eq:W}, that have the most suitable local models. This process is repeated until no further improvement is found. We empirically validate the advantage of using the escape heuristic in \autoref{app:escape}.

The pseudocode for \autoref{prob:add} (adding new data points to a {\sc slisemap} solution) is also given in \autoref{alg:slisemap} (function \texttt{Slisemap-new}). Here, we use the same escape heuristic to find a suitable neighbourhood as a starting point and then optimise the embedding and local model for the new data item(s) with PyTorch and L-BFGS.

The source code, published under an open source MIT license, as well as the code needed to replicate all of the experiments in this paper, is available via GitHub \citep{bjorklund2022slisemapgithub}.

\subsection{Computational complexity}
\label{sec:complexity}

Evaluation of the loss function of Equation \eqref{eq:main} requires at least $O(n^2m)$ iterations for linear regression and $O(n^2mp)$ for multinomial logistic regression. Because, for every local model $O(n)$, the prediction and loss $O(mp)$ must be calculated for every data item $O(n)$. The calculation of the soft neighbourhoods requires $O(n^2d)$ (from calculating the Euclidean distances), but $d<mp$ in most circumstances.

However, this is an iterative algorithm, where \autoref{eq:main} has to be evaluated multiple times. While it is difficult to provide strict running time limits for iterative optimisation algorithms such as L-BFGS---we study this experimentally in Section \ref{sec:experiments}---it is obvious that the algorithm may not scale well for very large ($n$) datasets.

However, usually it is sufficient to subsample $\min{(n,n_0)}$ data points, where $n_0$ is a suitably chosen constant, optimise for the loss function (\autoref{prob:slisemap}), and then add points to the existing solution (\autoref{prob:add}). By this procedure, the asymptotic complexity of {\sc slisemap} is linear with respect to the number of data points $n$. Especially for visualisation purposes, it often makes no sense to compute exact projection for a huge number of data points: visualisations cannot show more data points than there are pixels, so having an extremely accurate solution to the full optimisation problem instead of an approximate solution usually brings little additional benefit. Instead, finding a quick solution for sub-sampled data and adding the necessary number of data points to the embedding works well in practice, as shown in the experiments of Section \ref{sec:subset}.

\section{Experiments}
\label{sec:experiments}

In the experiments, we usually embed the data into two dimensions ($d=2$) and normalise data attributes, columns of the data matrix ${\bf X}$, to zero mean and unit variance as well as add an intercept term (column of ones) before running {\sc slisemap}. Furthermore, unless otherwise mentioned, we subsample the large datasets into $1000$ data items and run all experiments ten times.

Most datasets have been used in two scenarios, first as normal regression or classification using the definitions from \autoref{sec:regclass}, and second in an XAI-inspired scenario where the targets are predictions from black box models, using the alternative formulation from \autoref{sec:regclass} in the case of classification. When the white box model is a linear regression, we use $\lambda=10^{-4}$ as the regularisation coefficient and $\lambda=10^{-2}$ for logistic regression. An overview of the datasets and black box models can be seen in \autoref{tab:data}.

As explained earlier, we used PyTorch version 1.11 \citep{pytorch}. The runtime experiments were run on a server having an AMD Epyc processor at 2.4 GHz with 4 cores and 16 GB of memory allocated and an NVIDIA Tesla V100 GPU. The code to run the experiments is available via GitHub \citep{bjorklund2022slisemapgithub}.

\subsection{Datasets}
\label{sec:datasets}

In this section, we describe the datasets used in the experiments. The datasets and the black box models are available from \href{https://openml.org}{OpenML} \citep{vanschoren2014OpenML}. A quick summary can be seen in \autoref{tab:data}.

\begin{table}
	\centering
	\caption{An overview of the datasets and black box models used in the experiments.}
	\label{tab:data}
	\begin{tabular}{lrll}
		\toprule
		Dataset      & Size                & Task           & Black box model              \\
		\midrule
		{\sc rsynth} & $n \times m$        & Regression     & -                            \\
		Air Quality  & $7355 \times 11$    & Regression     & Random Forest                \\
		Boston       & $506 \times 13$     & Regression     & SVM                          \\
		Spam         & $4601 \times 57$    & Classification & Random Forest                \\
		Higgs        & $98~049 \times 28$  & Classification & Gradient Boosting            \\
		Covertype    & $581~011 \times 54$ & Classification & Logit Boost                  \\
		MNIST        & $70~000 \times 784$ & Classification & Convolutional Neural Network \\
		\bottomrule
	\end{tabular}
\end{table}

\paragraph{Synthetic data} We create synthetic {\em regression data} ({\sc rsynth}) as follows: given parameters dataset size $n$ (number of data items) and $m$ (number data attributes), as well as $k=3$ (number of clusters) and $s=0.25$ (standard deviation of the clusters). We first sample $j \in [k]$ coefficient vectors ${\bf \beta}_j\in{\mathbb{R}}^m$ from a normal distribution with zero mean and unit variance and cluster centroids ${\bf c}_j\in{\mathbb{R}}^m$ from a normal distribution with zero mean and standard deviation of $s$. We then create data items $i\in[n]$ by sampling the cluster index $j_i\in [k]$ uniformly and then generating a data vector ${\bf x}_i$ by sampling from a normal distribution with a mean of ${\bf c}_{j_i}$ and unit variance. The dependent variable is given by ${\bf y}_i={\bf x}_i^T{\bf \beta}_{j_i}+\epsilon_i$, where $\epsilon_i$ is Gaussian noise with zero mean and standard deviation of $0.1$.

\paragraph{Air Quality} data, cleaned and filtered as in
\cite{oikarinenDetectingVirtualConcept2021},
contains 7355 hourly instances of 12 different air quality measurements, one of which is used as a dependent variable and the others as covariates.

\paragraph{Boston Housing Dataset} collected by the U.S. census service from the Boston Standard Metropolitan Statistical Area in 1970. The size of the dataset is 506 items with 14 attributes, including the median value of owner-occupied homes that is used as the dependent variable.

\paragraph{Spam} \citep{cranor1998} is a UCI dataset containing both spam, i.e., unsolicited commercial email, as well as professional and personal emails. There are 4601 instances with 57 attributes (mostly word frequencies) in the dataset.

\paragraph{Higgs} \citep{baldi2014} is a UCI dataset containing 11 million simulated collision events for benchmarking classification algorithms. The dependent variable is whether a collision produces Higgs bosons. There are 28 attributes, the first 21 featuring kinematic properties measured by the particle detectors, and the last seven are functions of the first 21.

\paragraph{Covertype} is a UCI dataset with over half a million instances, used to classify forest cover type (seven different types, but we only use the first two) from 54 attributes. The areas represent natural forests with minimal human-caused disturbances.

\paragraph{MNIST} \citep{lecun1998gradientbased} is the classic machine learning dataset of handwritten digits from 0 to 9. Each digit is represented by a 28x28 greyscale image (784 pixels with integer pixel values between 0 and 255).
Due to the large number of pixels, we create a binary classification task by limiting the available digits to 2 and 3 and subsample them to 5000 data items.

\subsection{Metrics}
\label{sec:metrics}
To compare different {\sc slisemap} solutions, we want to be able to objectively measure the performance. To accomplish that, we consider the following metrics.

\paragraph{Loss} The most obvious thing to measure is the loss we are trying to minimise; see \autoref{eq:main}. However, the loss will change based on the parameters and the size of the dataset.

\paragraph{Cluster Purity} For the synthetic dataset, we know the ground truth, which means that we can compare the original clusters to the embedding found by {\sc slisemap}. If we denote the true cluster id:s as $c_1,\ldots,c_n$, we can measure how well low-dimensional embeddings reconstruct the true clusters:
\begin{equation}
	\frac{1}{n} \sum\nolimits_{i=1}^n |\textrm{k-NN}(i) \cap \{j \mid c_i = c_j \}| / k,
\end{equation}
where $\textrm{k-NN}(i)$ is the set of $k$ nearest neighbours (of item $i$) in the embedding space, using Euclidean distance, and $j \in [n]$. A larger value (closer to one) indicates that the dimensionality reduction has found the true clusters.

\paragraph{Fidelity}
The fidelity of a local model \citep{guidotti:2018:a} measures how well it can predict the correct outcome. Using the losses defined in \autoref{sec:regclass}, we obtain:
\begin{equation}
	\frac 1 n \sum\nolimits_{i=1}^n l(g_i({\bf{x}}_i), {\bf{y}}_i).
	\label{eq:fidelity1}
\end{equation}
We are interested not only in how the local models perform on the corresponding data items but also in how well they work for the neighbours in the embedding space, using, e.g., the $k$ nearest neighbours:
\begin{equation}
	\frac 1 n \sum\nolimits_{i=1}^n \frac 1 k \sum\nolimits_{j \in \textrm{k-NN}(i)} l(g_i({\bf{x}}_j), {\bf{y}}_j).
\end{equation}
A smaller value indicates better fidelity.

\paragraph{Coverage}
We also want local models that generalise to other data points. Otherwise, it would be trivial to find solutions. The coverage \citep{guidotti:2018:a} of a local model can be measured by counting the number of data items that have a loss less than a threshold $l_0$:
\begin{equation}
	\frac 1 n \sum\nolimits_{i=1}^n \frac 1 n \sum\nolimits_{j=1}^n (l(g_i({\bf{x}}_j), {\bf{y}}_j) < l_0).
	\label{eq:coverage1}
\end{equation}
This requires us to select the loss threshold $l_0$. Unless otherwise mentioned, in this paper, we choose the threshold to be the $0.3$ quantile of the losses of a global model (without the distance-based weights). Furthermore, we also want this behaviour to be reflected in the low-dimensional embedding. To verify this information, we limit the coverage testing to only the $k$ nearest neighbours:
\begin{equation}
	\frac 1 n \sum\nolimits_{i=1}^n \frac 1 k \sum\nolimits_{j \in \textrm{k-NN}(i)} (l(g_i({\bf{x}}_j), {\bf{y}}_j) < l_0).
\end{equation}
A larger coverage value (closer to one) is better.

\subsection{Parameter selection}
\label{sec:parameters}

{\sc slisemap} has one unusual parameter that needs to be selected: $z_{\rm radius}$. If $z_{\rm radius}$ is too small, then all data items are in the same cluster, resulting in underfitting local models that are almost identical to the global model. However, if $z_{\rm radius}$ is too large, then the neighbourhoods become singular, which causes the local models to overfit.

\begin{figure}
	\centering
	\includegraphics[width=\textwidth]{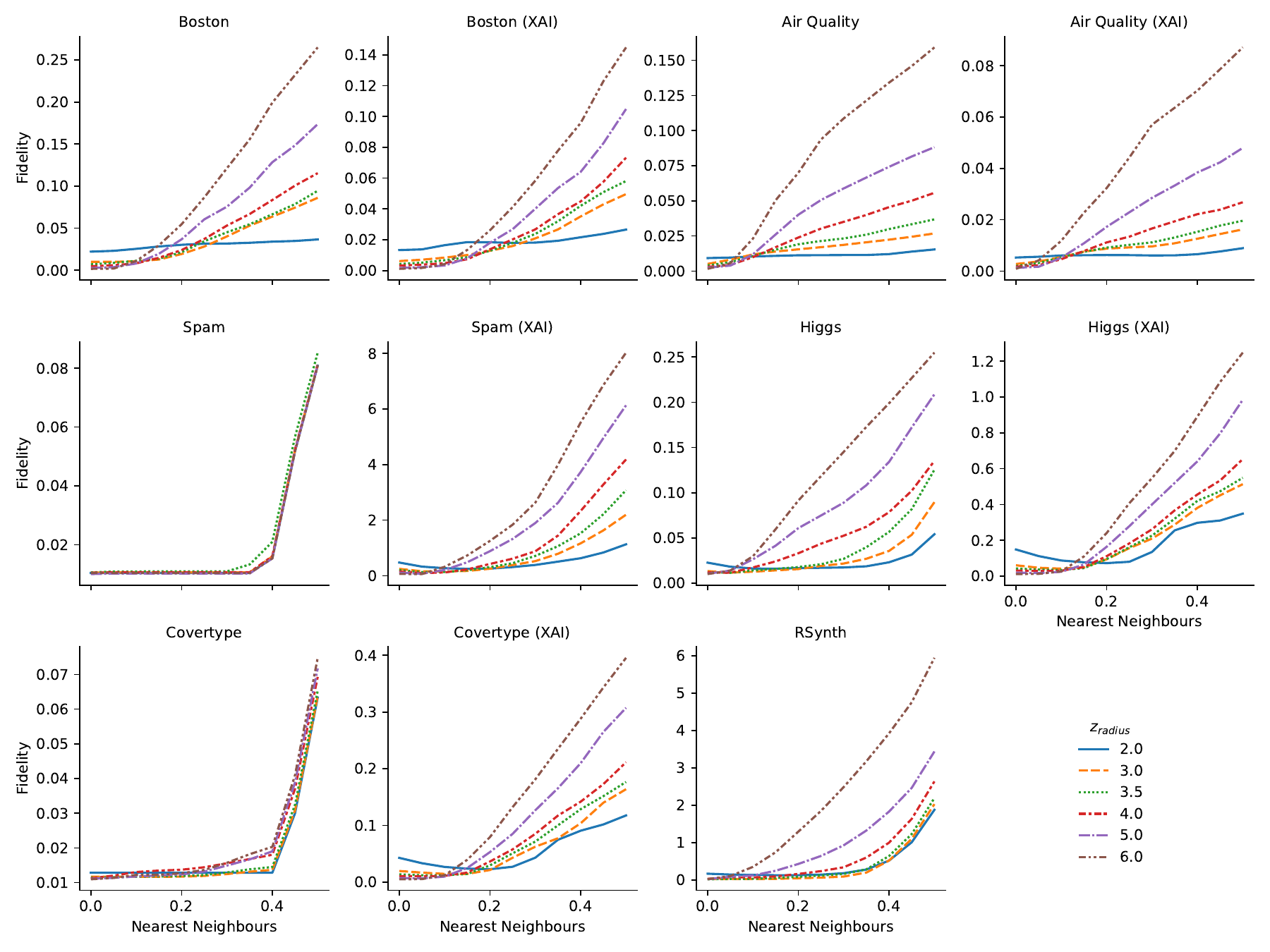}
	\caption{Fidelity of the local models versus the fraction of nearest neighbours (in the fidelity calculation) for different values of $z_{\rm radius}$. Smaller fidelity is better, especially for the nearest neighbors. Here, $3 \le z_{\rm radius} \le 4$ results in the best coverage.}
	\label{fig:param:fidelity}
\end{figure}

In \autoref{fig:param:fidelity}, we investigate how different values of $z_{\rm radius}$ affect the fidelity of the local models. Unless the local model is underfitting, the fidelity for the corresponding data item should be close to zero. Then, as the number of nearest neighbours grows, the fidelity should stay as low as possible for as long as possible to avoid overfitting. Based on these results, $z_{\rm radius}$ values from three to four seem to work well for all datasets.

We also consider how the coverage of the local models depends on the $z_{\rm radius}$. The coverage plots can be seen in \autoref{app:parameters} and support the same conclusion as the fidelity results. Thus, we use $z_{\rm radius} = 3.5$ as the default value for all the other experiments in this paper.

\subsection{Visualisations of the datasets}
\label{sec:vis}

\begin{figure}
	\centering\includegraphics[width=0.8\textwidth]{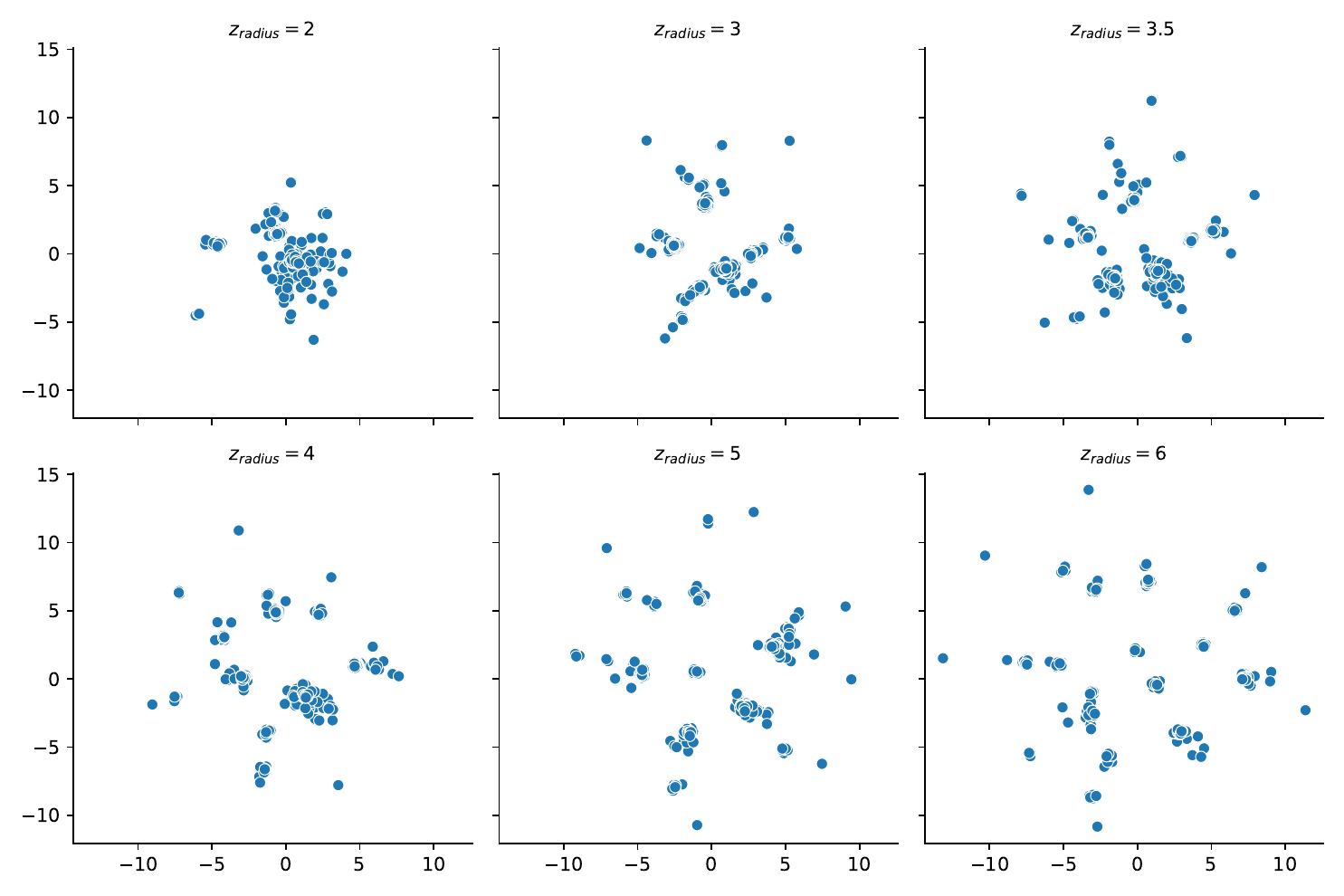}
	\caption{The low-dimensional embedding of the {\sc Boston} dataset with different values for $z_{\rm radius}$. Large values (bottom right) lead to sparse solutions with small clusters and potentially overfit local models. Small values of $z_{\rm radius}$ (top left) create a single dense cluster in the centre, with all (non-)local models being almost identical.}
	\label{fig:lineup}
\end{figure}

While fidelity and coverage can be used for the quantitative analysis of $z_{\rm radius}$, there is still room for a qualitative comparison to account for subjective preferences. In \autoref{fig:lineup}, we plot the low-dimensional embeddings for different $z_{\rm radius}$ values. At small values of $z_{\rm radius}$, all points converge to the same cluster, as expected. With large values, the points form smaller and smaller clusters, potentially leading to overfitting. Based on \autoref{fig:lineup}, $z_{\rm radius}$ values between three and four seem optimal, which matches the conclusions from above.

With {\sc slisemap}, we obtain not only an embedding but also local models for the data items. Data items that are nearby in the embedding space should have similar local models. We can verify this by clustering the local models independently of the embeddings and comparing these clusters to the structure in the embedding. Furthermore, models far apart in the embedding should look different due to the different local weights.

\begin{figure}
	\centering\includegraphics[width=0.8\textwidth]{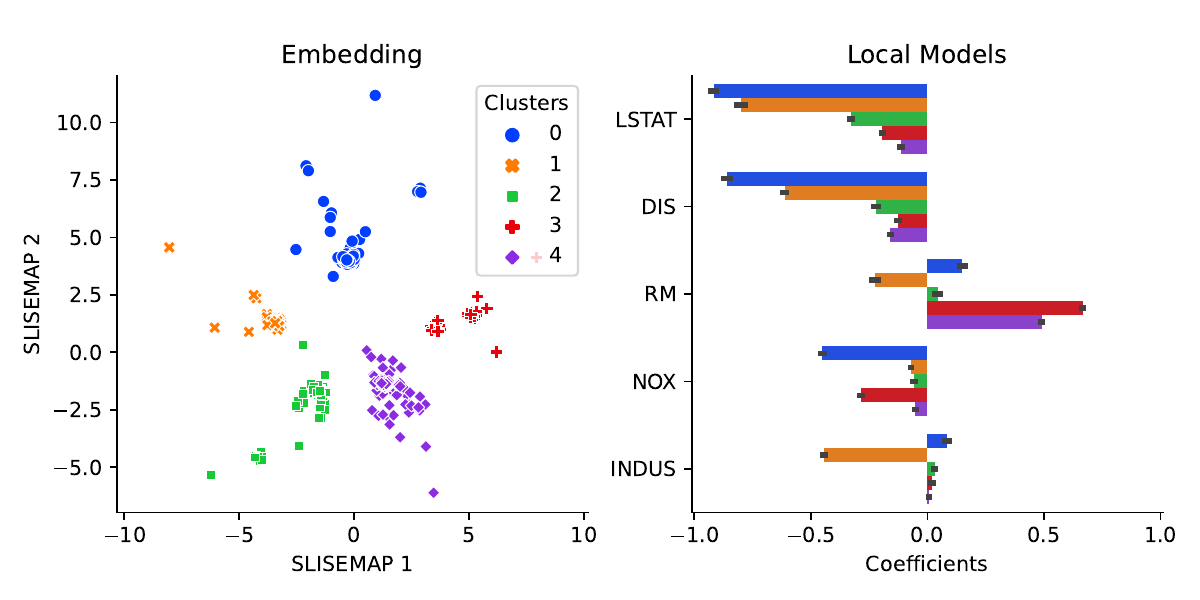}
	\caption{Clustering the local models (right) to see if they correspond to clusters in the embedding (left). This also shows that the clusters in the embedding (left) have distinct local models (right). The barplot (right) only shows the five most important attributes of the {\sc boston} dataset.}
	\label{fig:models}
\end{figure}

In \autoref{fig:models}, we cluster the coefficients of the local models using $k$-means clustering (on the {\sc boston} dataset). Here, we see that the clusters in the local models clearly match clusters in the embedding and that the clusters have different local models.
Looking at the local models, we notice something curious: industry (INDUS) generally has a weak positive coefficient, except for one cluster where it is strongly negative. Investigating this cluster reveals that this cluster has higher amounts of industry than the dataset in general (see density plots in \autoref{app:cluster_density}). Using {\sc slisemap}, we can learn that industry has a nonlinear effect on the value of homes; some industrial zoning might be good, but too much is detrimental.

A plot of the MNIST data is shown in \autoref{fig:mnist23} in the introduction, where we can see that some local models focus heavily on the bottom curve of 3:s, while others compare the differences between the pixels in the centre and the pixels just below the centre.

\subsection{Uniqueness}
\label{sec:unique}

In {\sc slisemap}, the embedding is influenced by the local models. Thus, if multiple local models are suitable for a particular data item, then the optimal embedding might be ambiguous. Some overlap between the local models is expected, and neighbouring (in the embedding) local models should be especially similar, due to the distance-based kernels in the loss function, \autoref{eq:main}. We also expect the hyperplanes of the local models to intersect, and any data items at these intersections will fit both models equally well.

\begin{figure}
	\centering
	\includegraphics[width=\textwidth]{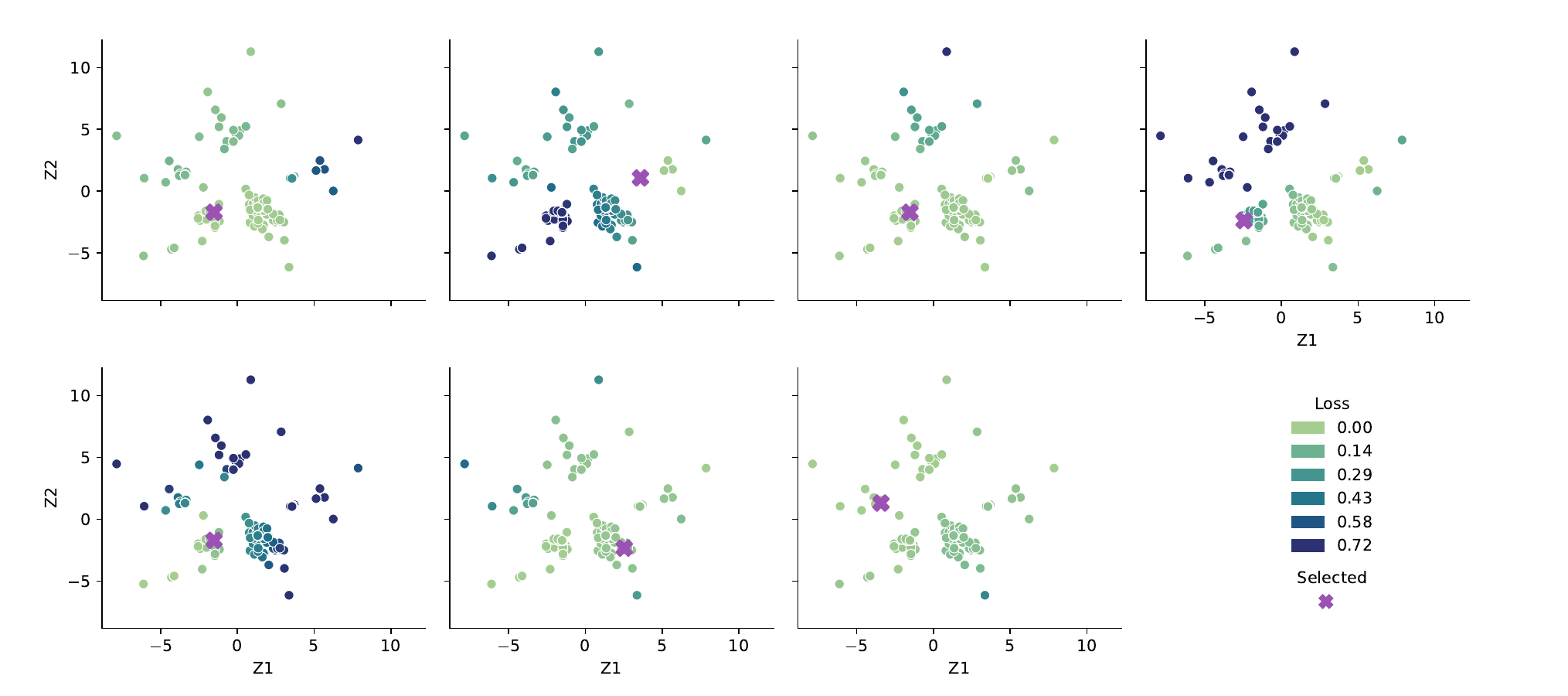}
	\caption{A {\sc slisemap} embedding for the {\sc boston} dataset. The embedding is plotted seven times with different data items selected. The points in the embedding are coloured based on how well the corresponding local model fits the selected data item. Some data items only fit local models that are nearby in the embedding (the same neighbourhood), while some data items are more general.}
	\label{fig:unique}
\end{figure}

In \autoref{fig:unique}, we select seven data items from the {\sc boston} dataset and plot scatterplots of the embedding, where the colour of each dot represents how suitable that local model is for the selected data item. We see that not all local models suit all data items, i.e., the local models are actually local. Furthermore, neighbouring points tend to have the most suitable local models, as expected. However, some data items fit well into multiple neighbourhoods.

These data items with multiple potential neighbourhoods make the solutions non-unique, since there are multiple local optima with almost equally good losses. However, as shown in \autoref{fig:models}, the local models in the different neighbourhoods are different, and this is important for the data items in \autoref{fig:unique} matching only a single neighbourhood.

\subsection{Subset sampling}
\label{sec:subset}

With large datasets, the quadratic scaling of {\sc slisemap}, see \autoref{sec:complexity}, can become problematic. One solution is to run {\sc slisemap} on a random subset of the data, and then, post hoc, add unseen data items whenever necessary (see \autoref{sec:adding}). With larger subsets, we expect better results, but with diminishing returns after the dataset is sufficiently covered.

To investigate how much data are needed, we randomly select $1000$ data items from the large datasets to be unseen test data and train {\sc slisemap} solutions on increasing numbers of data items sampled from the remaining data. Then, we add the unseen data items, using \slisemapnewf from \autoref{alg:slisemap}. We repeat this process ten times for each dataset and compare the fidelity, \autoref{eq:fidelity1}, between the training data and the test data.

\begin{figure}
	\centering
	\includegraphics[width=\textwidth]{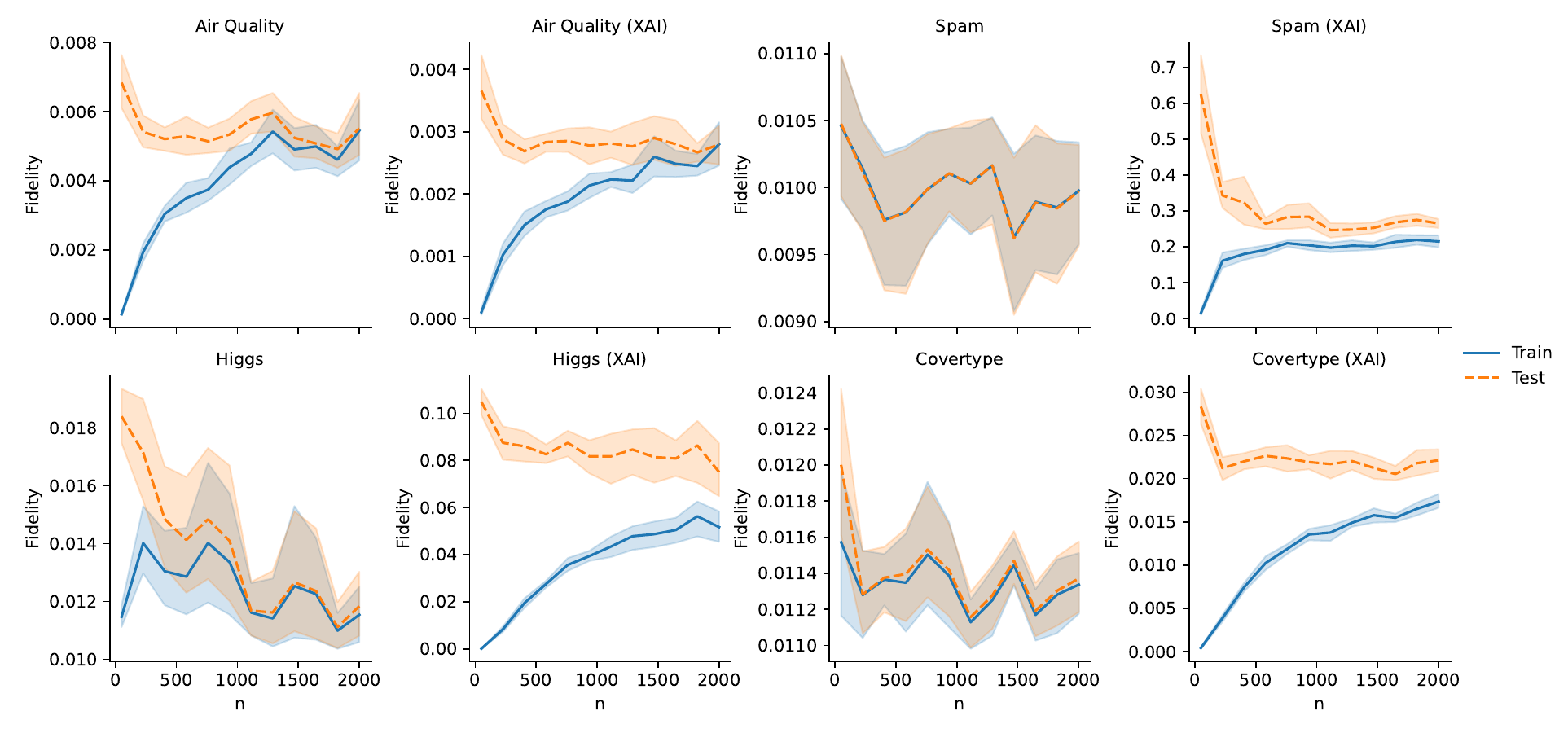}
	\caption{Adding new data items to {\sc slisemap} solutions trained on subsampled datasets. With a sufficiently large training dataset, the fidelity of unseen test data matches that of the training data. For most of these datasets, only a couple of hundreds of initial data items are required. Lower fidelity for the test data is better.}
	\label{fig:subset}
\end{figure}

The results can be seen in \autoref{fig:subset}. If the training dataset is too small, then the local models tend to overfit, but for most datasets, only a couple of hundred data items are needed to stabilise the results. This also coincides with the fidelities of the unseen test data approaching the fidelities of the training data.

\subsection{Higher dimensional embeddings}
\label{sec:dimensions}

In most experiments discussed in this paper, we use a two-dimensional embedding ($d=2$). This is because a two-dimensional embedding is easy to visualise, which we consider to be an important use-case for the embedding. However, {\sc slisemap} is not limited to only two dimensions, which we demonstrate in this section.

\begin{figure}
	\centering
	\includegraphics[width=\textwidth]{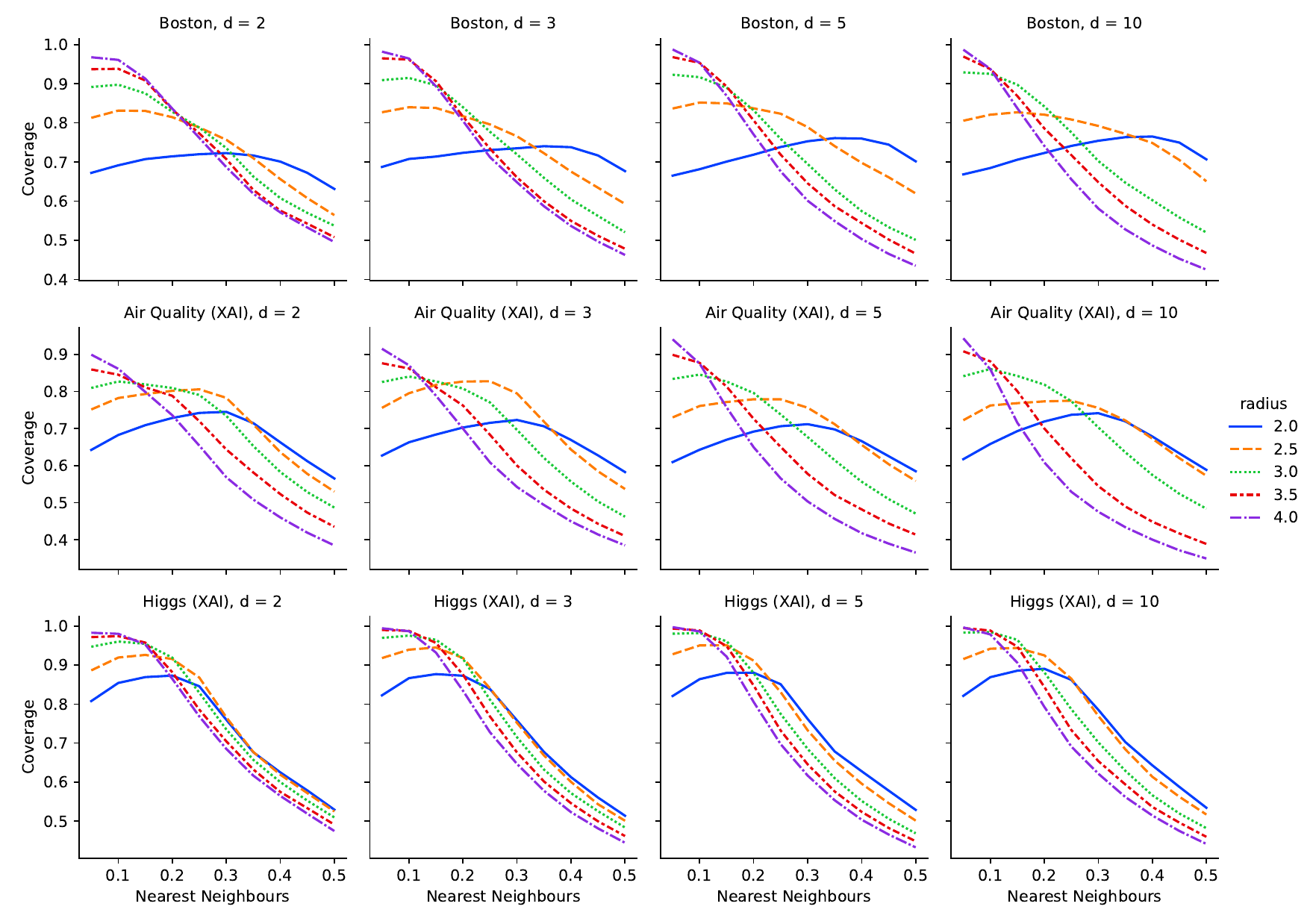}
	\caption{Coverage of the local models versus the fraction of nearest neighbours (in the coverage calculation) for different values of $z_{\rm radius}$ and different numbers of embedding dimensions $d$. As the threshold for the coverage, we use the $0.3$ quantile of the losses from a global model. Larger coverage is better, especially for the nearest neighbors. Here, $3 \le z_{\rm radius} \le 3.5$ results in the best coverage, even for higher dimensional embeddings. The full plot is available in \autoref{app:hdparams}.}
	\label{fig:hd:cov}
\end{figure}

Using the same fidelity and coverage metrics as in \autoref{sec:parameters}, we can find the best $z_{\rm radius}$ value for higher dimensional embeddings.
In \autoref{fig:hd:cov} and \autoref{app:hdparams}, we demonstrate that the same default parameter value of $z_{\rm radius} = 3.5$, which works well for two-dimensional embeddings, is also suitable for higher dimensions.

\begin{figure}
	\centering
	\includegraphics[width=\textwidth]{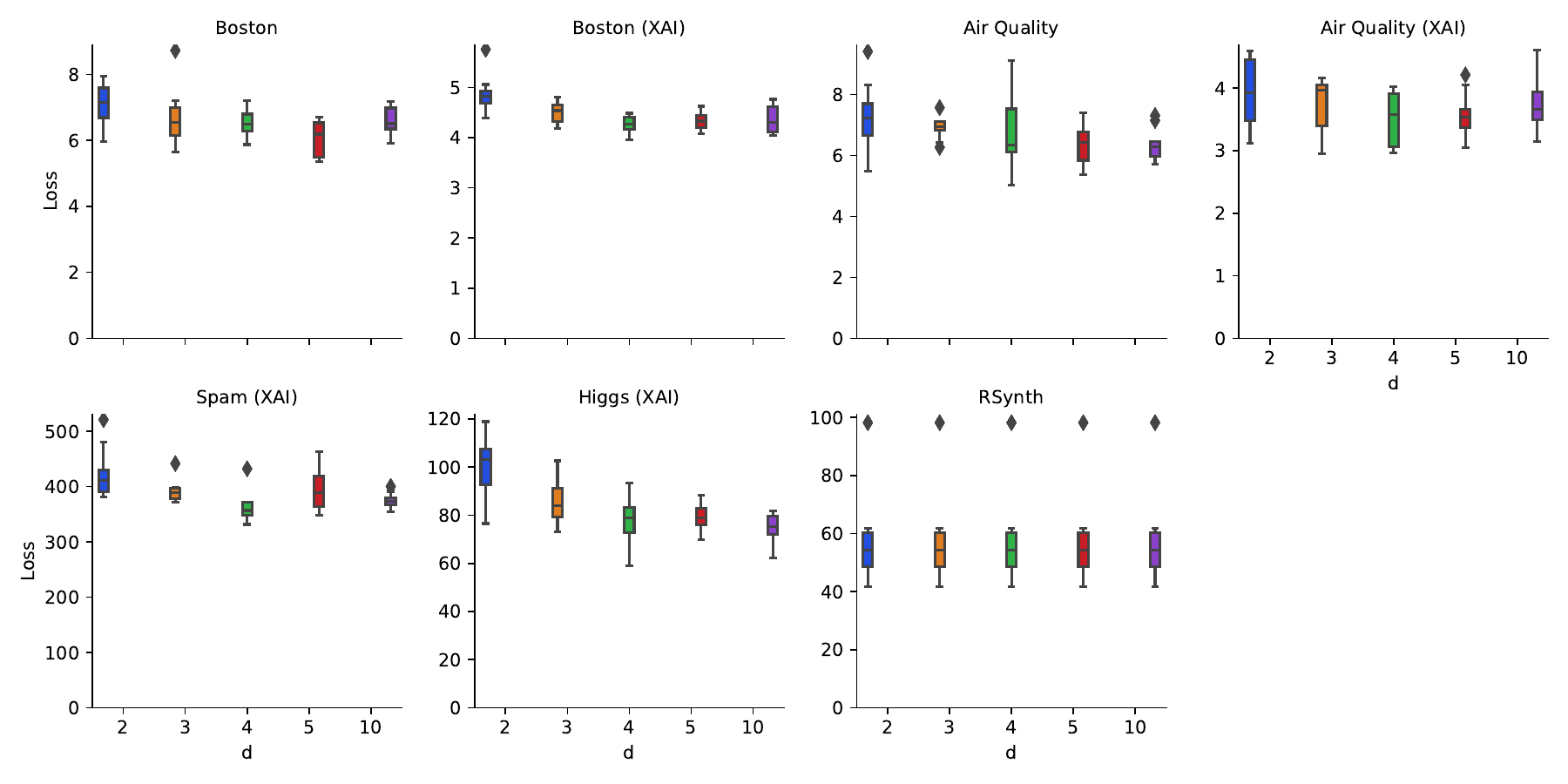}
	\caption{Comparing losses for different numbers of embedding dimensions $d$. With higher-dimensional embeddings, we expect either minor improvements to the loss due to more flexible distances between multiple clusters or that the loss stays roughly the same.}
	\label{fig:hd:loss}
\end{figure}

With two-dimensional embeddings, the intercluster distances are only independent for up to three clusters. This means that we expect higher dimensional embeddings to produce slightly lower losses if there are more than three clusters. In \autoref{fig:hd:loss}, we compare the losses for different numbers of dimensions. For some datasets, we indeed see minor improvements in the loss with increasing dimensionality. But, for example, in {\sc rsynth} we know that there are only three clusters, so higher dimensional embeddings offer no advantage. 

\subsection{GPU acceleration}
\label{sec:gpuexp}

\begin{figure}
	\centering
	\includegraphics[width=0.8\textwidth]{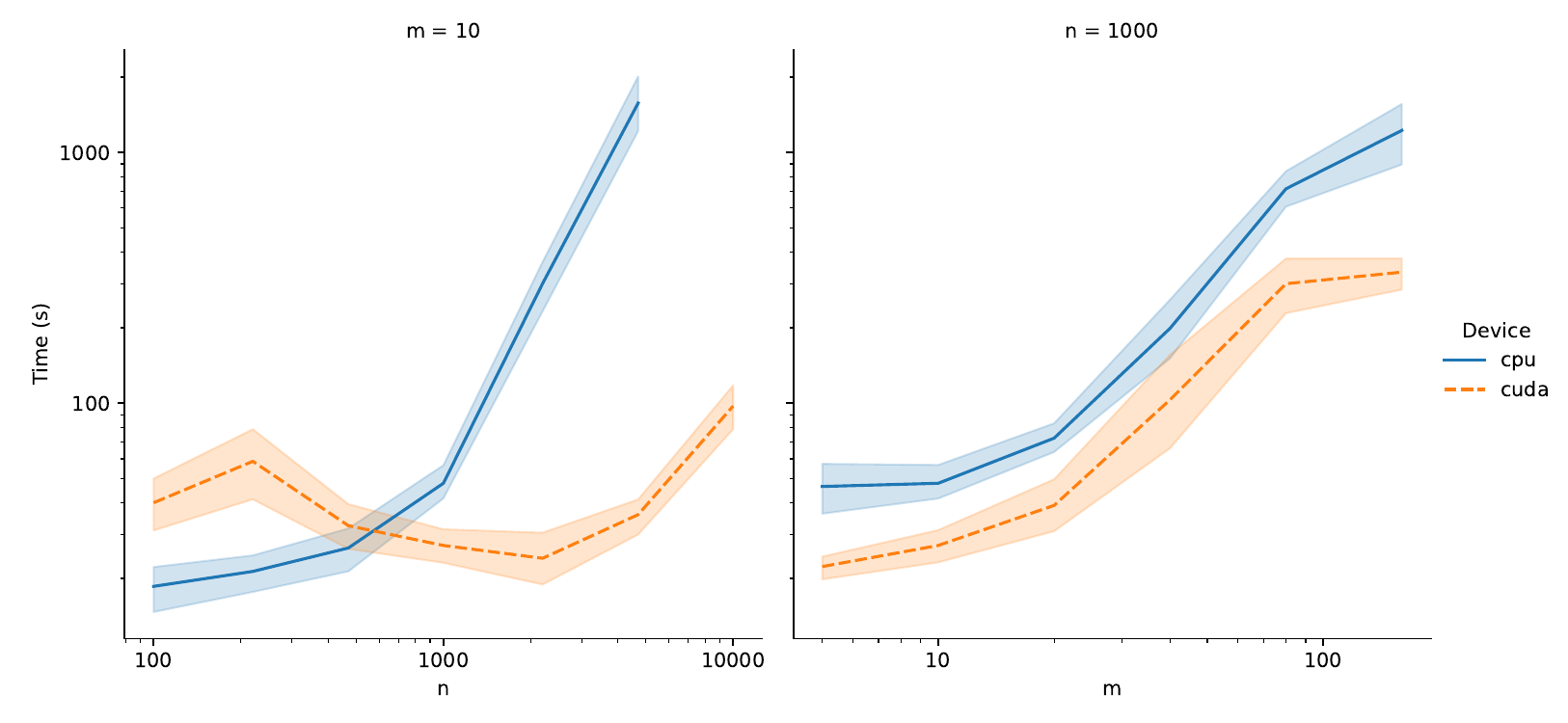}
	\caption{Runtimes for different dataset sizes (using the {\sc rsynth} dataset). GPU acceleration (cuda) brings some overhead but offers better parallelisation on large datasets. Note the logarithmic scale of the axis.}
	\label{fig:runrsynth}
\end{figure}

Since we implement {\sc slisemap} using PyTorch, the calculations can be accelerated using a GPU. Running {\sc slisemap} on a GPU should be faster than running on a CPU, especially for larger datasets. In \autoref{fig:runrsynth}, we apply {\sc slisemap} on {\sc rsynth} datasets with different sizes, both with and without GPU acceleration. The GPU implementation has some overhead, making it slower for small datasets (less than $400 \times 10$) but substantially faster for larger datasets.

\subsection{Comparison to dimensionality reduction methods}
\label{sec:cmp_dr}

The feature that differentiates {\sc slisemap} from other dimensionality reduction methods is that {\sc slisemap} provides both a low-dimensional embedding and local models. To demonstrate that doing this optimisation simultaneously is necessary, we take the embeddings from other dimensionality reduction methods and fit local models post hoc (essentially running {\sc slisemap} with a fixed ${\bf Z}$ given by the dimensionality reduction methods).

We use the following dimensionality reduction methods from the \texttt{Scikit-learn} package \citep{scikit-learn} for the comparison: PCA,
LLE \citep{roweis2000nonlinear}, MLLE \citep{zhang2007mlle}, MDS \citep{kruskal1964multidimensional}, ISOMAP \citep{isomap}, and t-SNE \citep{van2014accelerating}. We also consider UMAP \citep{2018arXivUMAP}.

A selection of the results can be seen in \autoref{tab:cmp_dr} (results for all datasets can be found in \autoref{app:dr}). Since none of the other methods consider the relationship between ${\bf X}$ and ${\bf y}$ (most do not even use ${\bf y}$), their post hoc local models are, unsurprisingly, nonoptimal. However, the downside of using {\sc slisemap} is the additional time required for convergence.

\begin{table}
	\centering
	\caption{Comparing {\sc slisemap} against other dimensionality reduction methods. The embeddings ${\bf Z}$ are given by the dimensionality reduction methods while the local model coefficients ${\bf B}$ are optimised post-hoc (using the {\sc slisemap} loss). Here we use $20 \%$ as the number of nearest neighbours, and the running times are without GPU-acceleration. The best results are in bold. The full table is available in \autoref{app:dr}.}
	\label{tab:cmp_dr}
	\begin{tabular}{llrrrrr}
		\toprule
		                &                    & Loss                    & Fidelity             & Fidelity NN          & Coverage NN          & Time (s)              \\
		Dataset         & Method             &                         &                      &                      &                      &                       \\
		\midrule
		\multicolumn{3}{l}{{\sc air quality}}                                                                                                                       \\
		$1000\times 11$ & Slisemap           & ${\bf7.87 \pm 1.33}$    & ${\bf0.00 \pm 0.00}$ & ${\bf0.02 \pm 0.01}$ & ${\bf0.73 \pm 0.06}$ & $160.20 \pm 54.18$    \\
		                & PCA                & $67.85 \pm 7.32$        & $0.05 \pm 0.01$      & $0.07 \pm 0.01$      & $0.33 \pm 0.01$      & $9.73 \pm 7.47$       \\
		                & Spectral Embedding & $74.81 \pm 8.34$        & $0.06 \pm 0.01$      & $0.07 \pm 0.01$      & $0.33 \pm 0.01$      & $8.16 \pm 7.63$       \\
		                & LLE                & $68.44 \pm 7.17$        & $0.05 \pm 0.01$      & $0.07 \pm 0.01$      & $0.33 \pm 0.02$      & $8.55 \pm 3.20$       \\
		                & MLLE               & $71.51 \pm 8.46$        & $0.06 \pm 0.01$      & $0.07 \pm 0.01$      & $0.33 \pm 0.01$      & $7.90 \pm 2.55$       \\
		                & MDS                & $68.08 \pm 7.59$        & $0.05 \pm 0.01$      & $0.07 \pm 0.01$      & $0.33 \pm 0.02$      & $22.88 \pm 9.23$      \\
		                & Non-Metric MDS     & $78.77 \pm 8.18$        & $0.06 \pm 0.01$      & $0.08 \pm 0.01$      & $0.30 \pm 0.02$      & ${\bf4.78 \pm 1.29}$  \\
		                & Isomap             & $67.83 \pm 7.64$        & $0.05 \pm 0.01$      & $0.07 \pm 0.01$      & $0.33 \pm 0.02$      & $7.81 \pm 2.69$       \\
		                & t-SNE              & $74.71 \pm 7.93$        & $0.06 \pm 0.01$      & $0.07 \pm 0.01$      & $0.33 \pm 0.01$      & $10.00 \pm 3.46$      \\
		                & UMAP               & $78.00 \pm 8.22$        & $0.07 \pm 0.01$      & $0.07 \pm 0.01$      & $0.32 \pm 0.01$      & $10.56 \pm 1.66$      \\
		                & Supervised UMAP    & $81.43 \pm 8.69$        & $0.08 \pm 0.01$      & $0.08 \pm 0.01$      & $0.31 \pm 0.01$      & ${\bf5.06 \pm 0.79}$  \\
		\multicolumn{3}{l}{{\sc spam (xai)}}                                                                                                                        \\
		$1000\times 57$ & Slisemap           & ${\bf492.45 \pm 66.51}$ & ${\bf0.20 \pm 0.03}$ & ${\bf0.31 \pm 0.07}$ & ${\bf0.86 \pm 0.04}$ & $237.68 \pm 76.62$    \\
		                & PCA                & $2700.08 \pm 100.82$    & $1.65 \pm 0.11$      & $2.31 \pm 0.11$      & $0.36 \pm 0.02$      & $18.35 \pm 4.87$      \\
		                & Spectral Embedding & $2522.60 \pm 82.28$     & $1.36 \pm 0.06$      & $2.07 \pm 0.07$      & $0.41 \pm 0.02$      & $17.48 \pm 4.36$      \\
		                & LLE                & $3075.05 \pm 233.23$    & $2.20 \pm 0.50$      & $3.09 \pm 0.41$      & $0.33 \pm 0.03$      & $19.22 \pm 5.55$      \\
		                & MLLE               & $3474.37 \pm 112.08$    & $3.15 \pm 0.13$      & $3.46 \pm 0.33$      & $0.29 \pm 0.02$      & ${\bf15.66 \pm 7.36}$ \\
		                & MDS                & $2401.03 \pm 39.46$     & $1.02 \pm 0.04$      & $2.11 \pm 0.07$      & $0.38 \pm 0.02$      & $77.63 \pm 15.93$     \\
		                & Non-Metric MDS     & $2926.17 \pm 85.26$     & $1.41 \pm 0.07$      & $2.70 \pm 0.09$      & $0.35 \pm 0.01$      & $24.51 \pm 4.44$      \\
		                & Isomap             & $2559.64 \pm 106.99$    & $1.38 \pm 0.12$      & $2.17 \pm 0.16$      & $0.38 \pm 0.02$      & $24.55 \pm 9.94$      \\
		                & t-SNE              & $2523.66 \pm 64.02$     & $1.22 \pm 0.04$      & $2.14 \pm 0.07$      & $0.39 \pm 0.02$      & $23.93 \pm 4.61$      \\
		                & UMAP               & $3352.64 \pm 190.68$    & $2.40 \pm 0.27$      & $2.92 \pm 0.32$      & $0.32 \pm 0.02$      & $19.96 \pm 3.51$      \\
		                & Supervised UMAP    & $3455.06 \pm 200.77$    & $2.63 \pm 0.31$      & $3.03 \pm 0.23$      & $0.31 \pm 0.02$      & ${\bf14.03 \pm 2.70}$ \\
		\multicolumn{3}{l}{{\sc rsynth}}                                                                                                                            \\
		$ 400\times 15$ & Slisemap           & ${\bf84.53 \pm 74.48}$  & ${\bf0.03 \pm 0.05}$ & ${\bf0.13 \pm 0.34}$ & ${\bf0.98 \pm 0.06}$ & $20.92 \pm 8.59$      \\
		                & PCA                & $2841.63 \pm 739.62$    & $3.23 \pm 0.81$      & $7.45 \pm 1.97$      & $0.37 \pm 0.02$      & ${\bf0.33 \pm 0.03}$  \\
		                & Spectral Embedding & $2921.21 \pm 836.71$    & $3.35 \pm 1.02$      & $7.62 \pm 2.19$      & $0.36 \pm 0.02$      & $0.39 \pm 0.02$       \\
		                & LLE                & $3151.40 \pm 933.08$    & $4.47 \pm 1.62$      & $8.58 \pm 2.41$      & $0.33 \pm 0.02$      & $0.53 \pm 0.06$       \\
		                & MLLE               & $3376.26 \pm 823.38$    & $7.04 \pm 1.40$      & $8.50 \pm 2.03$      & $0.33 \pm 0.01$      & $0.61 \pm 0.14$       \\
		                & MDS                & $2774.25 \pm 758.80$    & $2.89 \pm 0.78$      & $7.18 \pm 2.01$      & $0.37 \pm 0.02$      & $3.84 \pm 0.96$       \\
		                & Non-Metric MDS     & $3174.01 \pm 844.25$    & $3.54 \pm 0.93$      & $8.41 \pm 2.14$      & $0.33 \pm 0.02$      & $0.53 \pm 0.02$       \\
		                & Isomap             & $2938.62 \pm 860.11$    & $3.36 \pm 0.93$      & $7.73 \pm 2.36$      & $0.36 \pm 0.02$      & $0.45 \pm 0.04$       \\
		                & t-SNE              & $2987.75 \pm 815.80$    & $3.51 \pm 0.90$      & $7.71 \pm 2.03$      & $0.35 \pm 0.03$      & $1.19 \pm 0.02$       \\
		                & UMAP               & $3773.83 \pm 1086.93$   & $8.31 \pm 2.48$      & $8.75 \pm 2.54$      & $0.32 \pm 0.01$      & $5.62 \pm 0.20$       \\
		                & Supervised UMAP    & $3741.48 \pm 1058.81$   & $8.16 \pm 2.42$      & $8.66 \pm 2.42$      & $0.33 \pm 0.01$      & $2.28 \pm 0.02$       \\
		\bottomrule
	\end{tabular}
\end{table}

\subsection{Comparison to local explanation methods}
\label{sec:cmp_xai}
If we have access to a black box model, we can use {\sc slisemap} to find local and interpretable approximations of that black box model. In this section, we investigate how good the approximations are by checking both how local and how general the local models are. We also compare against other model-agnostic, local explanation methods. Furthermore, {\sc slisemap} finds all local models simultaneously, which could provide a speed benefit.

Of the local, model-agnostic, approximating explanations methods mentioned in \autoref{sec:related}, {\sc slisemap} is most closely related to {\sc slise} \citet{bjorklund2019}. {\sc slise} uses robust regression \citep{bjorklundRobustRegressionError2022} on data that have been centred on the selected data item to produce the local approximation. {\sc lime} \citep{ribeiro2016} creates a neighbourhood of synthetic data by mutating the selected data item (and using the black box model to obtain predictions). To increase interpretability {\sc lime}, normally, discretise continuous variables into binary variables (e.g., into quantiles). Then, {\sc lime} fits a least squares linear model to the synthetic neighbourhood to form the local approximation. {\sc shap} \citep{Lundberg_Lee_2017} tries to estimate the Shapley value of keeping a variable in the selected data item versus changing it. This is conceptually quite similar to the discretisation in {\sc lime}. The model-agnostic variants of {\sc shap} generally accomplishes this by creating variants of the selected data item where some of the variables are sampled from the dataset. These Shapley values are then used as the local approximation.

\begin{table}
	\centering
	\caption{Comparison of the local white box models given by {\sc slisemap}, {\sc slise}, {\sc shap}, {\sc lime}, and {\sc lime} with no discretisation. A global model is included as reference. The error tolerance for {\sc slise} and the coverage is selected such that the global model has a coverage of $0.3$, and the running times are without GPU-acceleration. Smaller fidelity and larger coverage are better.}
	\label{tab:cmp_xai}
	\begin{tabular}{llrrr}
		\toprule
		                &           & Fidelity             & Coverage             & Time (s)             \\
		Dataset         & Method    &                      &                      &                      \\
		\midrule
		{\sc boston (xai)}                                                                               \\
		$ 404\times 13$ & Slisemap  & $0.00 \pm 0.00$      & $0.35 \pm 0.02$      & $22.00 \pm 4.91$     \\
		                & SLISE     & ${\bf0.00 \pm 0.00}$ & ${\bf0.46 \pm 0.02}$ & $69.82 \pm 1.32$     \\
		                & SHAP      & ${\bf0.00 \pm 0.00}$ & $0.13 \pm 0.01$      & $169.20 \pm 3.65$    \\
		                & LIME      & $0.26 \pm 0.02$      & $0.14 \pm 0.01$      & $1614.38 \pm 20.60$  \\
		                & LIME (nd) & $0.16 \pm 0.01$      & $0.21 \pm 0.01$      & $49.89 \pm 0.73$     \\
		                & Global    & $0.11 \pm 0.01$      & $0.30 \pm 0.00$      & ${\bf0.01 \pm 0.00}$ \\
		{\sc air quality (xai)}                                                                          \\
		$1000\times 11$ & Slisemap  & $0.00 \pm 0.00$      & $0.26 \pm 0.01$      & $138.55 \pm 29.93$   \\
		                & SLISE     & ${\bf0.00 \pm 0.00}$ & ${\bf0.35 \pm 0.01}$ & $768.10 \pm 11.38$   \\
		                & SHAP      & $0.01 \pm 0.00$      & $0.08 \pm 0.00$      & $446.44 \pm 7.48$    \\
		                & LIME      & $0.23 \pm 0.05$      & $0.09 \pm 0.00$      & $3785.90 \pm 46.66$  \\
		                & LIME (nd) & $0.09 \pm 0.01$      & $0.26 \pm 0.02$      & $79.59 \pm 1.40$     \\
		                & Global    & $0.08 \pm 0.01$      & $0.30 \pm 0.00$      & ${\bf0.03 \pm 0.00}$ \\
		{\sc spam (xai)}                                                                                 \\
		$1000\times 57$ & Slisemap  & $0.21 \pm 0.01$      & $0.23 \pm 0.01$      & $228.32 \pm 101.68$  \\
		                & SLISE     & ${\bf0.00 \pm 0.00}$ & ${\bf0.57 \pm 0.01}$ & $578.25 \pm 66.37$   \\
		                & SHAP      & ${\bf0.00 \pm 0.00}$ & $0.15 \pm 0.01$      & $1891.84 \pm 18.94$  \\
		                & LIME      & $3.08 \pm 0.21$      & $0.20 \pm 0.01$      & $4445.11 \pm 80.06$  \\
		                & LIME (nd) & $9.56 \pm 0.27$      & $0.09 \pm 0.01$      & $200.16 \pm 0.97$    \\
		                & Global    & $3.10 \pm 0.12$      & $0.30 \pm 0.00$      & ${\bf0.35 \pm 0.16}$ \\
		{\sc higgs (xai)}                                                                                \\
		$1000\times 28$ & Slisemap  & $0.05 \pm 0.00$      & $0.28 \pm 0.01$      & $241.86 \pm 84.21$   \\
		                & SLISE     & ${\bf0.00 \pm 0.00}$ & ${\bf0.41 \pm 0.01}$ & $573.42 \pm 92.10$   \\
		                & SHAP      & ${\bf0.00 \pm 0.00}$ & $0.24 \pm 0.01$      & $512.41 \pm 20.85$   \\
		                & LIME      & $0.85 \pm 0.07$      & $0.27 \pm 0.01$      & $8983.48 \pm 236.47$ \\
		                & LIME (nd) & $1.34 \pm 0.05$      & $0.27 \pm 0.01$      & $161.86 \pm 4.08$    \\
		                & Global    & $1.19 \pm 0.06$      & $0.30 \pm 0.00$      & ${\bf0.15 \pm 0.37}$ \\
		\bottomrule
	\end{tabular}
\end{table}

In addition to the methods outlined above, {\sc slisemap}, {\sc slise}, {\sc shap}, and {\sc lime} (with and without discretisation), we also consider a global model as a reference. The global models allow us to check that the local approximations are indeed local (better fidelity than the global model) and how general the approximations are (by comparing the coverage). As the threshold for measuring coverage as well as the \emph{error tolerance} parameter in {\sc slise}, we use the $0.3$ quantile of the losses of the global model. The results can be seen in \autoref{tab:cmp_xai}.

By definition, {\sc slise} an {\sc shap} have perfect fidelity for the data item corresponding to the local model, with {\sc slisemap} not far behind. The global model is obviously not local and, thus, should have the worst fidelity. However, there is nothing in the {\sc lime} procedure that ensures that the local approximation matches the selected data item. This results in the fidelity of {\sc lime} being comparable to the global model.

One of the advantages of {\sc slise} is specifically optimising the subset size, which results in outstanding coverage. The local models in {\sc slisemap} are affected by the low-dimensional embedding. This reduced flexibility results in lower coverage than {\sc slise} but better coverage than both {\sc lime} and {\sc shap}.
Both {\sc shap} and {\sc lime} create synthetic neighbourhoods, which results in local models that are more difficult to generalise to real data items, reducing the coverage.

By computing all the local approximations at the same time, {\sc slisemap} tends to be faster than the methods doing it one-by-one, the exception being {\sc lime} with no discretisation. Furthermore, {\sc slisemap} also finds a low-dimensional embedding that can be used to visualise and compare different data items, different local approximations, and how they relate to each other.

\section{Conclusions}
\label{sec:discussion}

In this paper, we present a novel supervised manifold embedding method, {\sc slisemap}, that embeds data items into a lower-dimensional space such that nearby data items are modelled by the same white box model. Therefore, in addition to reducing the dimensionality of the data, {\sc slisemap} creates a visualisation that can be used to globally explore and explain black box classification and regression models.

We show that the state-of-the-art dimensionality reduction methods, unsurprisingly, cannot be used to explain classifiers or regression models. On the other hand, the state-of-the-art tools used to explain black box models typically only provide local explanations for single examples, whereas {\sc slisemap} gives an overview of all local explanations.

Interesting future work would be to explore how {\sc slisemap} visualisations can be used to better understand data, both with and without a black box model, and to help build better models. For example, if a {\sc slisemap} visualisation could show that some group of data items should be handled differently. Future work could also explore how to use {\sc slisemap} to detect anomalous behaviours, such as outliers or concept drift. Finally, the scaling of {\sc slisemap} could be improved by, e.g., using stochastic optimisation or prototypes.

The source code for {\sc slisemap}, published under an open source MIT license, as well as the code needed to replicate all of the experiments in this paper, is available via GitHub \citep{bjorklund2022slisemapgithub}.

\section*{Declarations}

{\bf Funding:}\hspace{0.1cm} Computational resources provided by \citeauthor{fcgi}.
~Anton Björklund is supported by the Doctoral Programme in Computer Science at University of Helsinki, and
Jarmo Mäkelä is supported by
Academy of Finland (decision 320182).\\
{\bf Conflict of interest:}\hspace{0.1cm} The authors declare that they have no conflicts of interest.\\
{\bf Ethical considerations:}\hspace{0.1cm} This paper is computational in nature, it uses only synthetic or previously published reference datasets, does not involve any human participants, and has no other outstanding ethical concerns.\\
{\bf Consent for participation and publication:}\hspace{0.1cm} Not applicable.\\
{\bf Data availability:}\hspace{0.1cm} All datasets found in this paper can be downloaded from \href{https://openml.org}{\url{openml.org}}.\\
{\bf Code availability:}\hspace{0.1cm} The source code for the algorithm and all the experiments are available under an open source MIT License from GitHub \citep{bjorklund2022slisemapgithub}.\\
{\bf Authors' contributions:}\hspace{0.1cm} The authors Anton Björklund, Jarmo Mäkelä, and Kai Puolamäki have all contributed to all parts of the research (theory, experiments, and writing).\\

\bibliographystyle{spbasic}      
\bibliography{ms}   
\label{end}

\pagebreak
\begin{appendix}

	\section{Additional parameter selection results}
	\label{app:parameters}

\begin{figure}[ht]
		\centering
		\includegraphics[width=\textwidth]{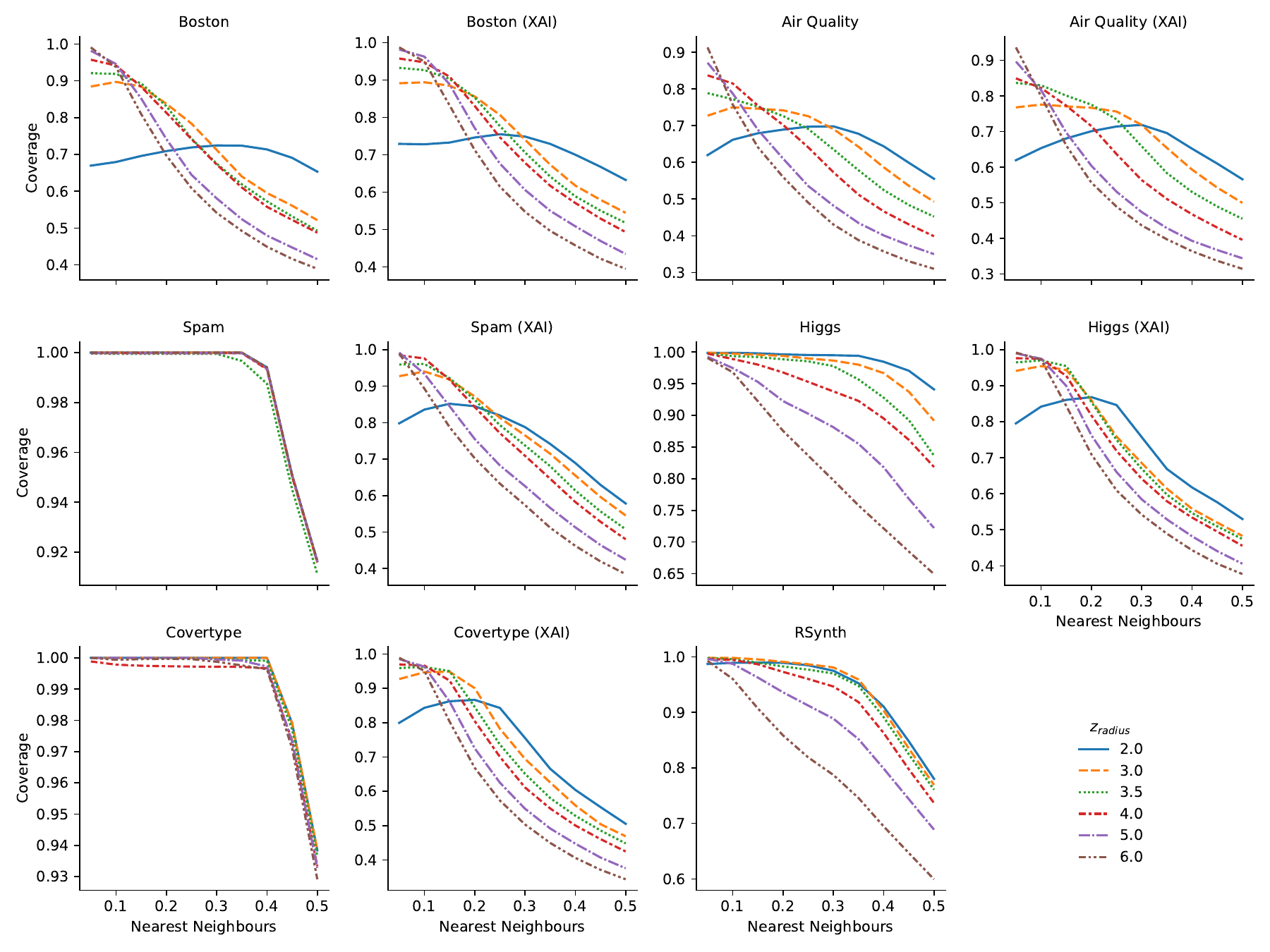}
		\caption{Coverage of the local models versus the fraction of nearest neighbours (in the coverage calculation) for different values for $z_{\rm radius}$. As the threshold for the coverage, we use the $0.3$ quantile of the losses from a global model. Larger coverage is better, especially for the nearest neighbors. Here, $3 \le z_{\rm radius} \le 4$ results in the best coverage.}
		\label{fig:param:coverage}
	\end{figure}

	In \autoref{fig:param:coverage}, we see how the coverage depends on the choice of value for $z_{\rm radius}$. The ideal value would be one where the coverage starts high and stays high as the number of nearest neighbours grows. If the coverage starts small, then we are probably underfitting, and if the coverage quickly drops, then we are probably overfitting. A $z_{\rm radius}$ value between three and four seems to be a good choice, which supports the conclusions in \autoref{sec:parameters}.

	\section{Escape Heuristic}
	\label{app:escape}

	In \autoref{sec:algo}, we describe a ``escape'' heuristic that we use to avoid getting stuck in a local optimum, which should yield better solutions. In \autoref{tab:escape_full}, we evaluate whether this is necessary. Using no heuristic would mean drastically faster running times. However, the solutions are nonoptimal compared to the full {\sc slisemap} solutions.

	\begin{table}[ht]
		\centering
		\caption{Comparing {\sc slisemap} with and without the escape heuristic. Not using the heuristic would be substantially faster, but result in much worse solutions. Here we use $20 \%$ as the number of nearest neighbours, and the best results are in bold.}
		\label{tab:escape_full}
		\begin{tabular}{llrrrrr}
			\toprule
							&           & Loss                    & Fidelity NN          & Coverage NN          & Cluster Purity       & Time (s)              \\
			Dataset         & Method    &                         &                      &                      &                      &                       \\
			\midrule
			\multicolumn{3}{l}{{\sc boston}}                                                                                                                   \\
			$ 404\times 13$ & Slisemap  & ${\bf7.91 \pm 0.80}$    & ${\bf0.02 \pm 0.01}$ & ${\bf0.83 \pm 0.03}$ &                      & $21.72 \pm 6.94$      \\
							& No escape & ${\bf7.65 \pm 0.55}$    & $0.05 \pm 0.01$      & $0.78 \pm 0.04$      &                      & ${\bf2.35 \pm 0.58}$  \\
			\multicolumn{3}{l}{{\sc boston (xai)}}                                                                                                             \\
			$ 404\times 13$ & Slisemap  & ${\bf5.42 \pm 0.33}$    & ${\bf0.01 \pm 0.00}$ & ${\bf0.85 \pm 0.02}$ &                      & $19.49 \pm 11.22$     \\
							& No escape & ${\bf5.62 \pm 0.53}$    & $0.03 \pm 0.01$      & $0.80 \pm 0.03$      &                      & ${\bf2.33 \pm 0.50}$  \\
			\multicolumn{3}{l}{{\sc air quality}}                                                                                                              \\
			$1000\times 11$ & Slisemap  & ${\bf7.87 \pm 1.33}$    & ${\bf0.02 \pm 0.01}$ & ${\bf0.73 \pm 0.06}$ &                      & $160.20 \pm 54.18$    \\
							& No escape & ${\bf7.38 \pm 0.75}$    & $0.03 \pm 0.00$      & ${\bf0.73 \pm 0.06}$ &                      & ${\bf13.88 \pm 2.15}$ \\
			\multicolumn{3}{l}{{\sc air quality (xai)}}                                                                                                        \\
			$1000\times 11$ & Slisemap  & ${\bf4.01 \pm 0.50}$    & ${\bf0.01 \pm 0.00}$ & ${\bf0.78 \pm 0.03}$ &                      & $162.91 \pm 39.20$    \\
							& No escape & ${\bf4.28 \pm 0.35}$    & $0.01 \pm 0.00$      & $0.75 \pm 0.05$      &                      & ${\bf11.24 \pm 3.31}$ \\
			\multicolumn{3}{l}{{\sc spam}}                                                                                                                     \\
			$1000\times 57$ & Slisemap  & ${\bf50.44 \pm 1.86}$   & ${\bf0.01 \pm 0.00}$ & ${\bf1.00 \pm 0.00}$ &                      & $96.51 \pm 18.59$     \\
							& No escape & $70.84 \pm 2.63$        & $0.05 \pm 0.01$      & $0.95 \pm 0.01$      &                      & ${\bf38.91 \pm 2.10}$ \\
			\multicolumn{3}{l}{{\sc spam (xai)}}                                                                                                               \\
			$1000\times 57$ & Slisemap  & ${\bf492.45 \pm 66.51}$ & ${\bf0.31 \pm 0.07}$ & ${\bf0.86 \pm 0.04}$ &                      & $237.68 \pm 76.62$    \\
							& No escape & $597.57 \pm 47.22$      & $0.65 \pm 0.20$      & ${\bf0.86 \pm 0.03}$ &                      & ${\bf16.48 \pm 3.36}$ \\
			\multicolumn{3}{l}{{\sc higgs}}                                                                                                                    \\
			$1000\times 28$ & Slisemap  & ${\bf53.31 \pm 2.79}$   & ${\bf0.02 \pm 0.00}$ & ${\bf0.99 \pm 0.01}$ &                      & $466.24 \pm 175.18$   \\
							& No escape & $218.88 \pm 85.72$      & $0.20 \pm 0.09$      & $0.56 \pm 0.28$      &                      & ${\bf43.81 \pm 4.55}$ \\
			\multicolumn{3}{l}{{\sc higgs (xai)}}                                                                                                              \\
			$1000\times 28$ & Slisemap  & ${\bf115.33 \pm 10.45}$ & ${\bf0.10 \pm 0.01}$ & ${\bf0.86 \pm 0.03}$ &                      & $304.91 \pm 132.32$   \\
							& No escape & $188.48 \pm 16.79$      & $0.37 \pm 0.05$      & $0.72 \pm 0.02$      &                      & ${\bf12.89 \pm 2.04}$ \\
			\multicolumn{3}{l}{{\sc covertype}}                                                                                                                \\
			$1000\times 54$ & Slisemap  & ${\bf54.02 \pm 0.80}$   & ${\bf0.01 \pm 0.00}$ & ${\bf1.00 \pm 0.00}$ &                      & $102.80 \pm 24.85$    \\
							& No escape & $66.83 \pm 1.69$        & $0.02 \pm 0.00$      & $0.98 \pm 0.01$      &                      & ${\bf37.82 \pm 2.25}$ \\
			\multicolumn{3}{l}{{\sc covertype (xai)}}                                                                                                          \\
			$1000\times 54$ & Slisemap  & ${\bf70.47 \pm 2.76}$   & ${\bf0.03 \pm 0.01}$ & ${\bf0.85 \pm 0.03}$ &                      & $402.65 \pm 143.24$   \\
							& No escape & $74.96 \pm 5.25$        & $0.06 \pm 0.02$      & ${\bf0.82 \pm 0.05}$ &                      & ${\bf19.24 \pm 2.84}$ \\
			\multicolumn{3}{l}{{\sc rsynth}}                                                                                                                   \\
			$ 400\times 15$ & Slisemap  & ${\bf84.53 \pm 74.48}$  & ${\bf0.13 \pm 0.34}$ & ${\bf0.98 \pm 0.06}$ & ${\bf0.89 \pm 0.12}$ & $20.92 \pm 8.59$      \\
							& No escape & $495.20 \pm 79.90$      & $2.53 \pm 0.58$      & $0.77 \pm 0.02$      & $0.38 \pm 0.02$      & ${\bf2.80 \pm 0.25}$  \\
			\bottomrule
		\end{tabular}
	\end{table}

	\section{Density plots for the clusters}
	\label{app:cluster_density}

\begin{figure}[ht]
		\centering
		\includegraphics[width=\textwidth]{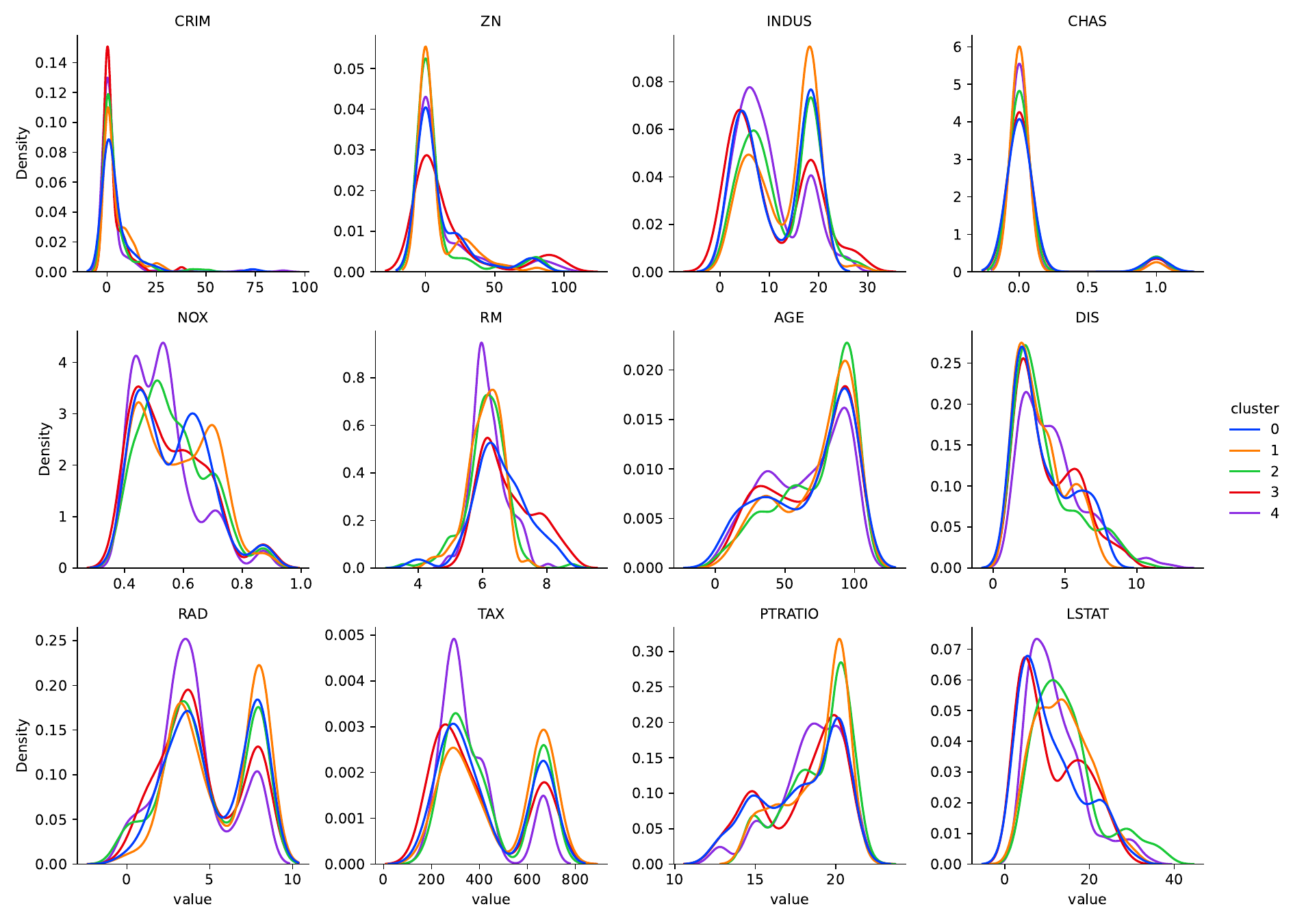}
		\caption{Density plots for the {\sc boston} dataset with clusters from the local models given by {\sc slisemap}.}
		\label{fig:cluster_density}
	\end{figure}

	In \autoref{sec:vis}, we qualitatively investigate a {\sc slisemap} solution for the {\sc boston} dataset. We find five clusters with different local models. To further study these clusters, we plot density plots for the clusters and variables in the dataset. The plots can be seen in \autoref{fig:cluster_density}.
	For example, we see that cluster 1 contains more industrial (INDUS) locations than average as well as better access to highways (RAD).

	\section{Higher-dimensional parameter selection results}
	\label{app:hdparams}

\begin{figure}[ht]
		\centering
		\includegraphics[width=\textwidth]{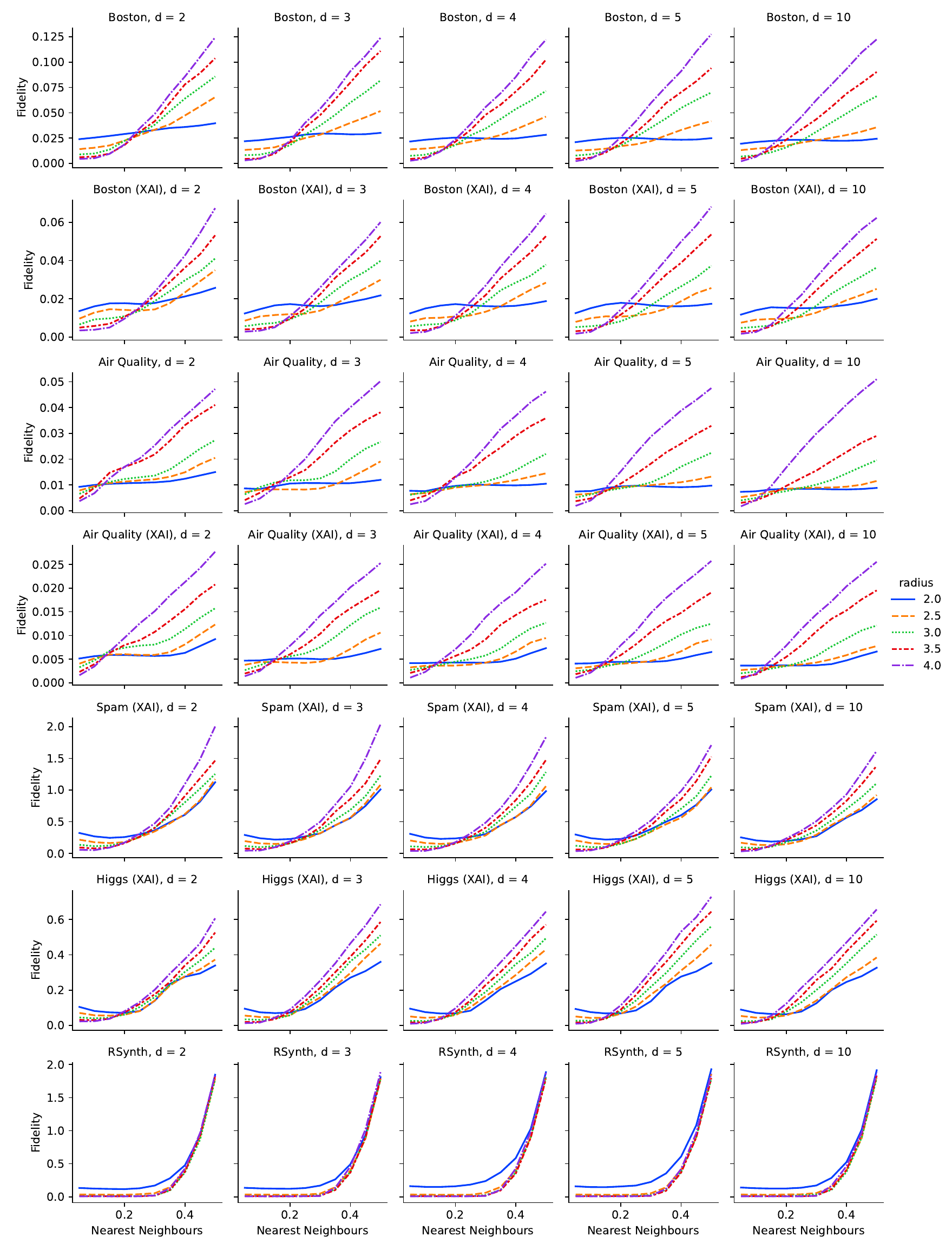}
		\caption{Fidelity of the local models versus the fraction of nearest neighbours (in the fidelity calculation) for different values of $z_{\rm radius}$ and different numbers of embedding dimensions $d$. Smaller fidelity is better, especially for the nearest neighbours. Here, $3 \le z_{\rm radius} \le 3.5$ results in the best fidelity, even for higher dimensional embeddings.}
		\label{fig:hd:fidelity}
	\end{figure}

\begin{figure}[ht]
		\centering
		\includegraphics[width=\textwidth]{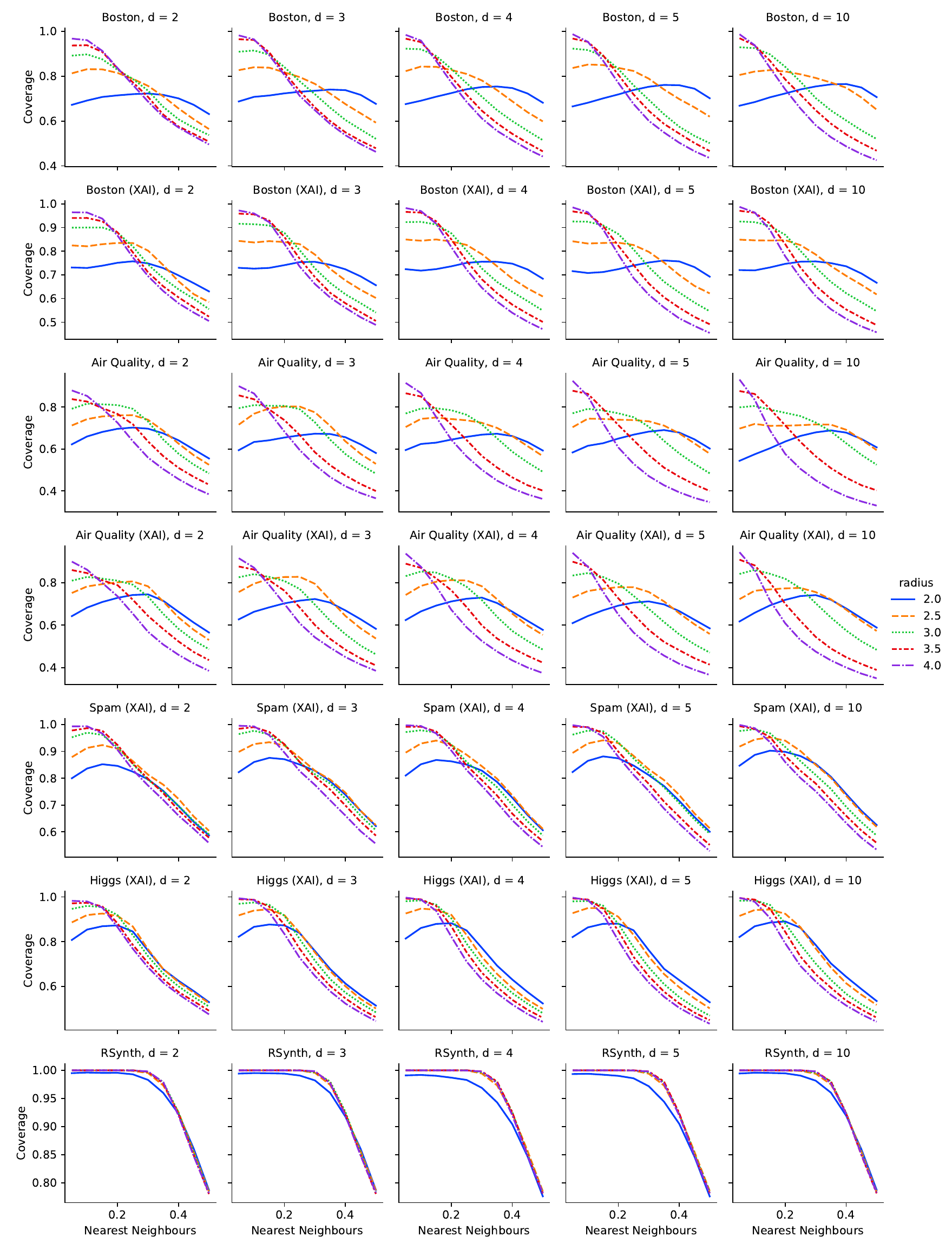}
		\caption{Coverage of the local models versus the fraction of nearest neighbours (in the coverage calculation) for different values for $z_{\rm radius}$ and different numbers of embedding dimensions $d$. As the threshold for coverage, we use the $0.3$ quantile of the losses from a global model. Larger coverage is better, especially for the nearest neighbours. Here, $3 \le z_{\rm radius} \le 3.5$ results in the best coverage, even for higher dimensional embeddings.}
		\label{fig:hd:coverage}
	\end{figure}

	When using {\sc slisemap} with embeddings of higher dimensions than two, in \autoref{sec:dimensions}, we need to select new values for the parameter $z_{\rm radius}$. For this, we employ the same procedure as in \autoref{sec:parameters} and \autoref{app:parameters}. The results for the fidelity can be seen in \autoref{fig:hd:fidelity}, and the results for coverage can be seen in \autoref{fig:hd:coverage}. These results support using $3.0 \le z_{\rm radius} \le 3.5$ for all datasets and different numbers of embedding dimensions. Thus, we use the same default value, $z_{\rm radius} = 3.5$, for higher dimensions as we do for two dimensions.

	\section{Additional dimensionality reduction results}
	\label{app:dr}

	In \autoref{sec:cmp_dr}, we compare {\sc slisemap} to other dimensionality reduction methods by post hoc training local models on the embeddings. In this appendix are additional comparisons to more datasets. We also include spectral embedding \citep{belkin2003laplacian}, nonmetric MDS \citep{kruskal1964multidimensional}, and supervised UMAP \citep{mcinnes2018umap-software} in the methods. In \autoref{tab:cmp_dr1}, we specifically investigate the synthetic dataset and find that {\sc slisemap} is the only method able to reconstruct the ground truth clusters. In Tables \ref{tab:cmp_dr2} and \ref{tab:cmp_dr3} are the full results for all the real datasets. In conclusion, {\sc slisemap} is the slowest of the methods but also the only one to provide reasonable local models.

	\begin{table}[ht]
		\centering
		\caption{Comparing {\sc slisemap} against other dimensionality reduction methods on the synthetic dataset. The embeddings ${\bf Z}$ are given by the dimensionality reduction methods while the local model coefficients ${\bf B}$ are optimised post-hoc (using the {\sc slisemap} loss). Here we use $20 \%$ as the number of nearest neighbours, and the running times are without GPU-acceleration. Note that this datasets is constructed such that you need to utilise ${\bf y}$ in order to find the known clusters. The best results are highlighted with bold.}
		\label{tab:cmp_dr1}
		\begin{tabular}{llrrrrrr}
			\toprule
							&                    & Loss                    & Fidelity             & Fidelity NN          & Coverage NN          & Cluster Purity       & Time (s)             \\
			Dataset         & Method             &                         &                      &                      &                      &                      &                      \\
			\midrule
			\multicolumn{3}{l}{{\sc rsynth}}                                                                                                                                                  \\
			$ 100\times  5$ & Slisemap           & ${\bf11.80 \pm 7.05}$   & ${\bf0.02 \pm 0.02}$ & ${\bf0.11 \pm 0.14}$ & ${\bf0.96 \pm 0.05}$ & ${\bf0.76 \pm 0.19}$ & $4.32 \pm 1.79$      \\
							& PCA                & $270.07 \pm 111.80$     & $1.22 \pm 0.51$      & $2.99 \pm 1.23$      & $0.41 \pm 0.06$      & $0.38 \pm 0.02$      & $0.37 \pm 0.03$      \\
							& Spectral Embedding & $294.93 \pm 124.35$     & $1.42 \pm 0.55$      & $3.15 \pm 1.45$      & $0.41 \pm 0.05$      & $0.37 \pm 0.02$      & $0.35 \pm 0.01$      \\
							& LLE                & $314.58 \pm 131.32$     & $1.75 \pm 0.80$      & $3.53 \pm 1.44$      & $0.38 \pm 0.05$      & $0.37 \pm 0.02$      & $0.46 \pm 0.05$      \\
							& MLLE               & $330.45 \pm 135.46$     & $2.38 \pm 0.90$      & $3.78 \pm 1.26$      & $0.35 \pm 0.05$      & $0.37 \pm 0.02$      & $0.52 \pm 0.09$      \\
							& MDS                & $275.08 \pm 113.00$     & $1.10 \pm 0.46$      & $3.03 \pm 1.26$      & $0.41 \pm 0.05$      & $0.38 \pm 0.02$      & $1.70 \pm 0.40$      \\
							& Non-Metric MDS     & $306.97 \pm 134.73$     & $1.08 \pm 0.50$      & $3.43 \pm 1.54$      & $0.39 \pm 0.07$      & $0.36 \pm 0.02$      & ${\bf0.33 \pm 0.01}$ \\
							& Isomap             & $290.58 \pm 121.49$     & $1.32 \pm 0.60$      & $3.28 \pm 1.41$      & $0.40 \pm 0.06$      & $0.37 \pm 0.02$      & $0.43 \pm 0.04$      \\
							& t-SNE              & $303.27 \pm 115.83$     & $1.51 \pm 0.56$      & $3.15 \pm 1.25$      & $0.39 \pm 0.05$      & $0.37 \pm 0.02$      & $0.56 \pm 0.01$      \\
							& UMAP               & $397.93 \pm 155.15$     & $3.34 \pm 1.32$      & $3.67 \pm 1.38$      & $0.33 \pm 0.04$      & $0.37 \pm 0.02$      & $5.52 \pm 0.09$      \\
							& Supervised UMAP    & $406.15 \pm 160.76$     & $3.56 \pm 1.46$      & $3.80 \pm 1.52$      & $0.32 \pm 0.02$      & $0.37 \pm 0.02$      & $2.04 \pm 0.02$      \\
			\multicolumn{3}{l}{{\sc rsynth}}                                                                                                                                                  \\
			$ 200\times 10$ & Slisemap           & ${\bf31.10 \pm 21.26}$  & ${\bf0.02 \pm 0.03}$ & ${\bf0.12 \pm 0.23}$ & ${\bf0.98 \pm 0.04}$ & ${\bf0.84 \pm 0.13}$ & $10.32 \pm 3.58$     \\
							& PCA                & $921.91 \pm 249.50$     & $1.82 \pm 0.52$      & $5.08 \pm 1.43$      & $0.34 \pm 0.02$      & $0.36 \pm 0.02$      & $0.37 \pm 0.04$      \\
							& Spectral Embedding & $982.53 \pm 288.10$     & $1.94 \pm 0.61$      & $5.29 \pm 1.59$      & $0.34 \pm 0.02$      & $0.37 \pm 0.02$      & ${\bf0.35 \pm 0.01}$ \\
							& LLE                & $1070.14 \pm 342.64$    & $3.13 \pm 1.20$      & $5.97 \pm 1.95$      & $0.33 \pm 0.02$      & $0.34 \pm 0.01$      & $0.50 \pm 0.04$      \\
							& MLLE               & $1091.79 \pm 309.72$    & $4.62 \pm 1.36$      & $5.74 \pm 1.58$      & $0.32 \pm 0.03$      & $0.35 \pm 0.00$      & $0.53 \pm 0.09$      \\
							& MDS                & $928.48 \pm 275.71$     & $1.64 \pm 0.51$      & $4.96 \pm 1.49$      & $0.35 \pm 0.02$      & $0.37 \pm 0.02$      & $2.33 \pm 0.61$      \\
							& Non-Metric MDS     & $1019.36 \pm 285.04$    & $1.76 \pm 0.42$      & $5.48 \pm 1.52$      & $0.32 \pm 0.03$      & $0.34 \pm 0.00$      & $0.42 \pm 0.01$      \\
							& Isomap             & $957.34 \pm 263.46$     & $1.86 \pm 0.56$      & $5.12 \pm 1.33$      & $0.34 \pm 0.02$      & $0.36 \pm 0.02$      & $0.42 \pm 0.03$      \\
							& t-SNE              & $989.12 \pm 289.49$     & $2.11 \pm 0.63$      & $5.16 \pm 1.51$      & $0.34 \pm 0.02$      & $0.35 \pm 0.01$      & $0.73 \pm 0.01$      \\
							& UMAP               & $1270.96 \pm 371.85$    & $5.46 \pm 1.62$      & $5.83 \pm 1.73$      & $0.32 \pm 0.02$      & $0.36 \pm 0.01$      & $5.81 \pm 0.34$      \\
							& Supervised UMAP    & $1281.29 \pm 371.56$    & $5.43 \pm 1.48$      & $5.92 \pm 1.64$      & $0.31 \pm 0.02$      & $0.36 \pm 0.01$      & $2.15 \pm 0.06$      \\
			\multicolumn{3}{l}{{\sc rsynth}}                                                                                                                                                  \\
			$ 400\times 15$ & Slisemap           & ${\bf84.53 \pm 74.48}$  & ${\bf0.03 \pm 0.05}$ & ${\bf0.13 \pm 0.34}$ & ${\bf0.98 \pm 0.06}$ & ${\bf0.89 \pm 0.12}$ & $20.92 \pm 8.59$     \\
							& PCA                & $2841.63 \pm 739.62$    & $3.23 \pm 0.81$      & $7.45 \pm 1.97$      & $0.37 \pm 0.02$      & $0.40 \pm 0.03$      & ${\bf0.33 \pm 0.03}$ \\
							& Spectral Embedding & $2921.21 \pm 836.71$    & $3.35 \pm 1.02$      & $7.62 \pm 2.19$      & $0.36 \pm 0.02$      & $0.40 \pm 0.02$      & $0.39 \pm 0.02$      \\
							& LLE                & $3151.40 \pm 933.08$    & $4.47 \pm 1.62$      & $8.58 \pm 2.41$      & $0.33 \pm 0.02$      & $0.35 \pm 0.01$      & $0.53 \pm 0.06$      \\
							& MLLE               & $3376.26 \pm 823.38$    & $7.04 \pm 1.40$      & $8.50 \pm 2.03$      & $0.33 \pm 0.01$      & $0.35 \pm 0.01$      & $0.61 \pm 0.14$      \\
							& MDS                & $2774.25 \pm 758.80$    & $2.89 \pm 0.78$      & $7.18 \pm 2.01$      & $0.37 \pm 0.02$      & $0.40 \pm 0.03$      & $3.84 \pm 0.96$      \\
							& Non-Metric MDS     & $3174.01 \pm 844.25$    & $3.54 \pm 0.93$      & $8.41 \pm 2.14$      & $0.33 \pm 0.02$      & $0.34 \pm 0.00$      & $0.53 \pm 0.02$      \\
							& Isomap             & $2938.62 \pm 860.11$    & $3.36 \pm 0.93$      & $7.73 \pm 2.36$      & $0.36 \pm 0.02$      & $0.39 \pm 0.02$      & $0.45 \pm 0.04$      \\
							& t-SNE              & $2987.75 \pm 815.80$    & $3.51 \pm 0.90$      & $7.71 \pm 2.03$      & $0.35 \pm 0.03$      & $0.37 \pm 0.02$      & $1.19 \pm 0.02$      \\
							& UMAP               & $3773.83 \pm 1086.93$   & $8.31 \pm 2.48$      & $8.75 \pm 2.54$      & $0.32 \pm 0.01$      & $0.37 \pm 0.01$      & $5.62 \pm 0.20$      \\
							& Supervised UMAP    & $3741.48 \pm 1058.81$   & $8.16 \pm 2.42$      & $8.66 \pm 2.42$      & $0.33 \pm 0.01$      & $0.38 \pm 0.01$      & $2.28 \pm 0.02$      \\
			\multicolumn{3}{l}{{\sc rsynth}}                                                                                                                                                  \\
			$ 800\times 20$ & Slisemap           & ${\bf174.25 \pm 29.87}$ & ${\bf0.01 \pm 0.00}$ & ${\bf0.03 \pm 0.03}$ & ${\bf1.00 \pm 0.00}$ & ${\bf0.94 \pm 0.01}$ & $49.74 \pm 16.97$    \\
							& PCA                & $8298.25 \pm 1542.92$   & $5.46 \pm 1.08$      & $10.58 \pm 2.00$     & $0.38 \pm 0.03$      & $0.45 \pm 0.04$      & ${\bf1.31 \pm 0.94}$ \\
							& Spectral Embedding & $8472.20 \pm 1491.20$   & $5.62 \pm 1.03$      & $10.77 \pm 1.98$     & $0.39 \pm 0.03$      & $0.45 \pm 0.04$      & ${\bf1.10 \pm 0.22}$ \\
							& LLE                & $9589.32 \pm 1613.10$   & $7.26 \pm 1.26$      & $12.80 \pm 2.21$     & $0.31 \pm 0.01$      & $0.34 \pm 0.00$      & $1.50 \pm 0.64$      \\
							& MLLE               & $10487.73 \pm 1975.77$  & $11.96 \pm 2.52$     & $12.77 \pm 2.25$     & $0.31 \pm 0.01$      & $0.35 \pm 0.01$      & $1.43 \pm 0.36$      \\
							& MDS                & $8399.70 \pm 1632.89$   & $5.45 \pm 1.15$      & $10.64 \pm 2.19$     & $0.38 \pm 0.03$      & $0.43 \pm 0.03$      & $16.86 \pm 5.92$     \\
							& Non-Metric MDS     & $10054.49 \pm 1621.43$  & $7.08 \pm 1.13$      & $13.13 \pm 2.10$     & $0.31 \pm 0.01$      & $0.34 \pm 0.00$      & $2.24 \pm 0.28$      \\
							& Isomap             & $9045.78 \pm 1601.37$   & $6.14 \pm 1.10$      & $11.69 \pm 2.18$     & $0.35 \pm 0.03$      & $0.42 \pm 0.04$      & ${\bf1.28 \pm 0.39}$ \\
							& t-SNE              & $9198.38 \pm 1625.66$   & $6.44 \pm 1.18$      & $11.76 \pm 2.08$     & $0.35 \pm 0.02$      & $0.38 \pm 0.03$      & $3.69 \pm 0.84$      \\
							& UMAP               & $10941.08 \pm 1855.82$  & $12.04 \pm 2.13$     & $12.73 \pm 2.24$     & $0.31 \pm 0.01$      & $0.41 \pm 0.03$      & $8.32 \pm 1.06$      \\
							& Supervised UMAP    & $10829.11 \pm 1857.73$  & $11.60 \pm 2.13$     & $12.47 \pm 2.21$     & $0.32 \pm 0.01$      & $0.42 \pm 0.03$      & $4.39 \pm 0.71$      \\
			\multicolumn{3}{l}{{\sc rsynth}}                                                                                                                                                  \\
			$1000\times 25$ & Slisemap           & ${\bf252.41 \pm 43.03}$ & ${\bf0.01 \pm 0.00}$ & ${\bf0.03 \pm 0.03}$ & ${\bf1.00 \pm 0.00}$ & ${\bf0.94 \pm 0.01}$ & $77.29 \pm 50.34$    \\
							& PCA                & $12141.25 \pm 1919.31$  & $6.32 \pm 1.13$      & $12.29 \pm 1.95$     & $0.40 \pm 0.03$      & $0.47 \pm 0.02$      & ${\bf1.09 \pm 0.14}$ \\
							& Spectral Embedding & $12335.77 \pm 2183.05$  & $6.50 \pm 1.30$      & $12.50 \pm 2.41$     & $0.40 \pm 0.04$      & $0.46 \pm 0.03$      & $1.36 \pm 0.15$      \\
							& LLE                & $14291.92 \pm 2287.89$  & $8.87 \pm 1.38$      & $15.20 \pm 2.42$     & $0.32 \pm 0.01$      & $0.34 \pm 0.00$      & $2.11 \pm 0.17$      \\
							& MLLE               & $15422.12 \pm 2321.97$  & $14.00 \pm 2.06$     & $14.98 \pm 2.20$     & $0.32 \pm 0.01$      & $0.35 \pm 0.01$      & $2.02 \pm 0.54$      \\
							& MDS                & $12278.10 \pm 1808.29$  & $6.42 \pm 0.95$      & $12.34 \pm 1.86$     & $0.38 \pm 0.03$      & $0.42 \pm 0.03$      & $24.06 \pm 4.50$     \\
							& Non-Metric MDS     & $14677.33 \pm 2251.64$  & $8.26 \pm 1.26$      & $15.34 \pm 2.33$     & $0.32 \pm 0.01$      & $0.34 \pm 0.00$      & $3.97 \pm 2.36$      \\
							& Isomap             & $13354.91 \pm 2031.37$  & $7.30 \pm 1.27$      & $13.80 \pm 2.17$     & $0.36 \pm 0.03$      & $0.41 \pm 0.03$      & $1.58 \pm 0.12$      \\
							& t-SNE              & $12902.31 \pm 1934.43$  & $6.96 \pm 1.05$      & $13.12 \pm 2.05$     & $0.36 \pm 0.03$      & $0.40 \pm 0.03$      & $4.23 \pm 0.55$      \\
							& UMAP               & $16117.32 \pm 2418.49$  & $14.14 \pm 2.23$     & $14.91 \pm 2.25$     & $0.32 \pm 0.01$      & $0.42 \pm 0.02$      & $9.30 \pm 2.04$      \\
							& Supervised UMAP    & $16041.85 \pm 2333.70$  & $13.71 \pm 1.91$     & $14.69 \pm 2.17$     & $0.32 \pm 0.01$      & $0.43 \pm 0.02$      & $4.68 \pm 0.35$      \\
			\bottomrule
		\end{tabular}
	\end{table}
	\begin{table}[ht]
		\centering
		\caption{Comparing {\sc slisemap} against other dimensionality reduction methods on real datasets. The embeddings ${\bf Z}$ are given by the dimensionality reduction methods while the local model coefficients ${\bf B}$ are optimised post-hoc (using the {\sc slisemap} loss). Here we use $20 \%$ as the number of nearest neighbours, and the running times are without GPU-acceleration. The best results are highlighted with bold, continued in \autoref{tab:cmp_dr3}.}
		\label{tab:cmp_dr2}
		\begin{tabular}{llrrrrr}
			\toprule
							&                    & Loss                  & Fidelity             & Fidelity NN          & Coverage NN          & Time (s)               \\
			Dataset         & Method             &                       &                      &                      &                      &                        \\
			\midrule
			\multicolumn{3}{l}{{\sc boston}}                                                                                                                           \\
			$ 404\times 13$ & Slisemap           & ${\bf7.91 \pm 0.80}$  & ${\bf0.01 \pm 0.00}$ & ${\bf0.02 \pm 0.01}$ & ${\bf0.83 \pm 0.03}$ & $21.72 \pm 6.94$       \\
							& PCA                & $58.24 \pm 3.17$      & $0.08 \pm 0.00$      & $0.13 \pm 0.01$      & $0.40 \pm 0.02$      & $1.64 \pm 0.16$        \\
							& Spectral Embedding & $59.97 \pm 3.89$      & $0.11 \pm 0.01$      & $0.14 \pm 0.01$      & $0.37 \pm 0.02$      & $1.04 \pm 0.14$        \\
							& LLE                & $63.25 \pm 4.55$      & $0.14 \pm 0.01$      & $0.16 \pm 0.01$      & $0.36 \pm 0.02$      & $0.71 \pm 0.13$        \\
							& MLLE               & $59.17 \pm 5.08$      & $0.13 \pm 0.01$      & $0.26 \pm 0.10$      & $0.37 \pm 0.02$      & $1.36 \pm 0.33$        \\
							& MDS                & $54.70 \pm 3.46$      & $0.07 \pm 0.00$      & $0.13 \pm 0.01$      & $0.41 \pm 0.02$      & $5.22 \pm 0.78$        \\
							& Non-Metric MDS     & $86.09 \pm 6.04$      & $0.10 \pm 0.01$      & $0.22 \pm 0.02$      & $0.30 \pm 0.01$      & ${\bf0.65 \pm 0.04}$   \\
							& Isomap             & $53.76 \pm 3.31$      & $0.09 \pm 0.01$      & $0.14 \pm 0.01$      & $0.38 \pm 0.03$      & $1.56 \pm 0.26$        \\
							& t-SNE              & $60.68 \pm 3.83$      & $0.09 \pm 0.01$      & $0.15 \pm 0.01$      & $0.39 \pm 0.02$      & $2.58 \pm 0.30$        \\
							& UMAP               & $66.06 \pm 3.99$      & $0.12 \pm 0.01$      & $0.15 \pm 0.01$      & $0.38 \pm 0.03$      & $6.54 \pm 0.57$        \\
							& Supervised UMAP    & $66.88 \pm 4.41$      & $0.13 \pm 0.01$      & $0.15 \pm 0.01$      & $0.39 \pm 0.02$      & $2.97 \pm 0.30$        \\
			\multicolumn{3}{l}{{\sc boston (xai)}}                                                                                                                     \\
			$ 404\times 13$ & Slisemap           & ${\bf5.42 \pm 0.33}$  & ${\bf0.00 \pm 0.00}$ & ${\bf0.01 \pm 0.00}$ & ${\bf0.85 \pm 0.02}$ & $19.49 \pm 11.22$      \\
							& PCA                & $33.39 \pm 2.07$      & $0.04 \pm 0.00$      & $0.07 \pm 0.01$      & $0.48 \pm 0.03$      & $1.38 \pm 0.08$        \\
							& Spectral Embedding & $40.28 \pm 1.45$      & $0.07 \pm 0.00$      & $0.09 \pm 0.00$      & $0.42 \pm 0.02$      & $1.08 \pm 0.10$        \\
							& LLE                & $47.78 \pm 7.07$      & $0.09 \pm 0.02$      & $0.12 \pm 0.01$      & $0.40 \pm 0.02$      & $0.80 \pm 0.30$        \\
							& MLLE               & $34.92 \pm 4.15$      & $0.07 \pm 0.01$      & $0.15 \pm 0.04$      & $0.41 \pm 0.02$      & $1.25 \pm 0.19$        \\
							& MDS                & $30.00 \pm 1.22$      & $0.03 \pm 0.00$      & $0.06 \pm 0.00$      & $0.50 \pm 0.01$      & $5.08 \pm 1.11$        \\
							& Non-Metric MDS     & $69.88 \pm 3.34$      & $0.07 \pm 0.00$      & $0.18 \pm 0.01$      & $0.30 \pm 0.01$      & ${\bf0.64 \pm 0.05}$   \\
							& Isomap             & $32.04 \pm 2.08$      & $0.04 \pm 0.00$      & $0.09 \pm 0.01$      & $0.46 \pm 0.02$      & $1.44 \pm 0.25$        \\
							& t-SNE              & $40.15 \pm 1.21$      & $0.05 \pm 0.00$      & $0.09 \pm 0.00$      & $0.43 \pm 0.01$      & $2.34 \pm 0.20$        \\
							& UMAP               & $43.83 \pm 3.94$      & $0.07 \pm 0.00$      & $0.10 \pm 0.01$      & $0.43 \pm 0.02$      & $6.42 \pm 0.40$        \\
							& Supervised UMAP    & $45.10 \pm 5.36$      & $0.08 \pm 0.01$      & $0.10 \pm 0.01$      & $0.42 \pm 0.02$      & $2.79 \pm 0.16$        \\
			\multicolumn{3}{l}{{\sc air quality}}                                                                                                                      \\
			$1000\times 11$ & Slisemap           & ${\bf7.87 \pm 1.33}$  & ${\bf0.00 \pm 0.00}$ & ${\bf0.02 \pm 0.01}$ & ${\bf0.73 \pm 0.06}$ & $160.20 \pm 54.18$     \\
							& PCA                & $67.85 \pm 7.32$      & $0.05 \pm 0.01$      & $0.07 \pm 0.01$      & $0.33 \pm 0.01$      & $9.73 \pm 7.47$        \\
							& Spectral Embedding & $74.81 \pm 8.34$      & $0.06 \pm 0.01$      & $0.07 \pm 0.01$      & $0.33 \pm 0.01$      & $8.16 \pm 7.63$        \\
							& LLE                & $68.44 \pm 7.17$      & $0.05 \pm 0.01$      & $0.07 \pm 0.01$      & $0.33 \pm 0.02$      & $8.55 \pm 3.20$        \\
							& MLLE               & $71.51 \pm 8.46$      & $0.06 \pm 0.01$      & $0.07 \pm 0.01$      & $0.33 \pm 0.01$      & $7.90 \pm 2.55$        \\
							& MDS                & $68.08 \pm 7.59$      & $0.05 \pm 0.01$      & $0.07 \pm 0.01$      & $0.33 \pm 0.02$      & $22.88 \pm 9.23$       \\
							& Non-Metric MDS     & $78.77 \pm 8.18$      & $0.06 \pm 0.01$      & $0.08 \pm 0.01$      & $0.30 \pm 0.02$      & ${\bf4.78 \pm 1.29}$   \\
							& Isomap             & $67.83 \pm 7.64$      & $0.05 \pm 0.01$      & $0.07 \pm 0.01$      & $0.33 \pm 0.02$      & $7.81 \pm 2.69$        \\
							& t-SNE              & $74.71 \pm 7.93$      & $0.06 \pm 0.01$      & $0.07 \pm 0.01$      & $0.33 \pm 0.01$      & $10.00 \pm 3.46$       \\
							& UMAP               & $78.00 \pm 8.22$      & $0.07 \pm 0.01$      & $0.07 \pm 0.01$      & $0.32 \pm 0.01$      & $10.56 \pm 1.66$       \\
							& Supervised UMAP    & $81.43 \pm 8.69$      & $0.08 \pm 0.01$      & $0.08 \pm 0.01$      & $0.31 \pm 0.01$      & ${\bf5.06 \pm 0.79}$   \\
			\multicolumn{3}{l}{{\sc air quality (xai)}}                                                                                                                \\
			$1000\times 11$ & Slisemap           & ${\bf4.01 \pm 0.50}$  & ${\bf0.00 \pm 0.00}$ & ${\bf0.01 \pm 0.00}$ & ${\bf0.78 \pm 0.03}$ & $162.91 \pm 39.20$     \\
							& PCA                & $35.22 \pm 3.63$      & $0.03 \pm 0.00$      & $0.03 \pm 0.00$      & $0.35 \pm 0.01$      & $7.23 \pm 1.26$        \\
							& Spectral Embedding & $38.99 \pm 4.66$      & $0.03 \pm 0.00$      & $0.04 \pm 0.00$      & $0.35 \pm 0.01$      & ${\bf5.63 \pm 1.20}$   \\
							& LLE                & $36.59 \pm 4.19$      & $0.03 \pm 0.00$      & $0.04 \pm 0.00$      & $0.34 \pm 0.01$      & ${\bf6.55 \pm 2.07}$   \\
							& MLLE               & $37.62 \pm 4.28$      & $0.03 \pm 0.00$      & $0.04 \pm 0.00$      & $0.35 \pm 0.01$      & $7.67 \pm 3.13$        \\
							& MDS                & $35.29 \pm 3.72$      & $0.03 \pm 0.00$      & $0.03 \pm 0.00$      & $0.36 \pm 0.01$      & $33.94 \pm 12.78$      \\
							& Non-Metric MDS     & $41.21 \pm 5.40$      & $0.03 \pm 0.00$      & $0.04 \pm 0.01$      & $0.30 \pm 0.01$      & $7.34 \pm 2.54$        \\
							& Isomap             & $35.20 \pm 3.68$      & $0.03 \pm 0.00$      & $0.03 \pm 0.00$      & $0.35 \pm 0.01$      & $9.43 \pm 2.52$        \\
							& t-SNE              & $38.67 \pm 4.58$      & $0.03 \pm 0.00$      & $0.04 \pm 0.00$      & $0.35 \pm 0.01$      & $8.97 \pm 2.78$        \\
							& UMAP               & $41.23 \pm 5.28$      & $0.04 \pm 0.00$      & $0.04 \pm 0.00$      & $0.34 \pm 0.01$      & $11.03 \pm 1.66$       \\
							& Supervised UMAP    & $41.29 \pm 5.34$      & $0.04 \pm 0.00$      & $0.04 \pm 0.00$      & $0.33 \pm 0.01$      & ${\bf6.01 \pm 0.83}$   \\
			\multicolumn{3}{l}{{\sc spam}}                                                                                                                             \\
			$1000\times 57$ & Slisemap           & ${\bf50.44 \pm 1.86}$ & ${\bf0.01 \pm 0.00}$ & ${\bf0.01 \pm 0.00}$ & ${\bf1.00 \pm 0.00}$ & $96.51 \pm 18.59$      \\
							& PCA                & $224.82 \pm 4.00$     & $0.17 \pm 0.01$      & $0.19 \pm 0.01$      & $0.69 \pm 0.02$      & $52.99 \pm 9.21$       \\
							& Spectral Embedding & $207.47 \pm 4.51$     & $0.16 \pm 0.00$      & $0.18 \pm 0.01$      & $0.71 \pm 0.02$      & $56.67 \pm 10.12$      \\
							& LLE                & $273.74 \pm 7.99$     & $0.25 \pm 0.02$      & $0.27 \pm 0.01$      & $0.35 \pm 0.17$      & ${\bf30.16 \pm 17.48}$ \\
							& MLLE               & $283.45 \pm 1.53$     & $0.28 \pm 0.00$      & $0.28 \pm 0.00$      & $0.15 \pm 0.12$      & ${\bf19.29 \pm 18.12}$ \\
							& MDS                & $233.36 \pm 2.76$     & $0.17 \pm 0.01$      & $0.20 \pm 0.01$      & $0.63 \pm 0.02$      & $104.83 \pm 41.92$     \\
							& Non-Metric MDS     & $283.90 \pm 1.60$     & $0.27 \pm 0.00$      & $0.28 \pm 0.00$      & $0.34 \pm 0.02$      & ${\bf22.15 \pm 18.49}$ \\
							& Isomap             & $213.29 \pm 5.39$     & $0.16 \pm 0.01$      & $0.18 \pm 0.01$      & $0.69 \pm 0.02$      & $46.82 \pm 14.44$      \\
							& t-SNE              & $207.71 \pm 4.65$     & $0.15 \pm 0.01$      & $0.18 \pm 0.01$      & $0.68 \pm 0.01$      & $46.75 \pm 7.73$       \\
							& UMAP               & $256.87 \pm 10.13$    & $0.20 \pm 0.01$      & $0.22 \pm 0.01$      & $0.53 \pm 0.06$      & $49.26 \pm 13.75$      \\
							& Supervised UMAP    & $69.71 \pm 23.99$     & $0.02 \pm 0.00$      & $0.02 \pm 0.01$      & $0.99 \pm 0.01$      & $52.50 \pm 13.28$      \\
			\bottomrule
		\end{tabular}
	\end{table}
	\begin{table}[ht]
		\centering
		\caption{Comparison of {\sc slisemap} with other dimensionality reduction methods on real datasets. The embeddings ${\bf Z}$ are given by the dimensionality reduction methods while the local model coefficients ${\bf B}$ are optimised post-hoc (using the {\sc slisemap} loss). Here we use $20 \%$ as the number of nearest neighbours, and the running times are without GPU-acceleration. The best results are highlighted with bold, continued from \autoref{tab:cmp_dr2}.}
		\label{tab:cmp_dr3}
		\begin{tabular}{llrrrrr}
			\toprule
							&                    & Loss                    & Fidelity             & Fidelity NN          & Coverage NN          & Time (s)               \\
			Dataset         & Method             &                         &                      &                      &                      &                        \\
			\midrule
			\multicolumn{3}{l}{{\sc spam (xai)}}                                                                                                                         \\
			$1000\times 57$ & Slisemap           & ${\bf492.45 \pm 66.51}$ & ${\bf0.20 \pm 0.03}$ & ${\bf0.31 \pm 0.07}$ & ${\bf0.86 \pm 0.04}$ & $237.68 \pm 76.62$     \\
							& PCA                & $2700.08 \pm 100.82$    & $1.65 \pm 0.11$      & $2.31 \pm 0.11$      & $0.36 \pm 0.02$      & $18.35 \pm 4.87$       \\
							& Spectral Embedding & $2522.60 \pm 82.28$     & $1.36 \pm 0.06$      & $2.07 \pm 0.07$      & $0.41 \pm 0.02$      & $17.48 \pm 4.36$       \\
							& LLE                & $3075.05 \pm 233.23$    & $2.20 \pm 0.50$      & $3.09 \pm 0.41$      & $0.33 \pm 0.03$      & $19.22 \pm 5.55$       \\
							& MLLE               & $3474.37 \pm 112.08$    & $3.15 \pm 0.13$      & $3.46 \pm 0.33$      & $0.29 \pm 0.02$      & ${\bf15.66 \pm 7.36}$  \\
							& MDS                & $2401.03 \pm 39.46$     & $1.02 \pm 0.04$      & $2.11 \pm 0.07$      & $0.38 \pm 0.02$      & $77.63 \pm 15.93$      \\
							& Non-Metric MDS     & $2926.17 \pm 85.26$     & $1.41 \pm 0.07$      & $2.70 \pm 0.09$      & $0.35 \pm 0.01$      & $24.51 \pm 4.44$       \\
							& Isomap             & $2559.64 \pm 106.99$    & $1.38 \pm 0.12$      & $2.17 \pm 0.16$      & $0.38 \pm 0.02$      & $24.55 \pm 9.94$       \\
							& t-SNE              & $2523.66 \pm 64.02$     & $1.22 \pm 0.04$      & $2.14 \pm 0.07$      & $0.39 \pm 0.02$      & $23.93 \pm 4.61$       \\
							& UMAP               & $3352.64 \pm 190.68$    & $2.40 \pm 0.27$      & $2.92 \pm 0.32$      & $0.32 \pm 0.02$      & $19.96 \pm 3.51$       \\
							& Supervised UMAP    & $3455.06 \pm 200.77$    & $2.63 \pm 0.31$      & $3.03 \pm 0.23$      & $0.31 \pm 0.02$      & ${\bf14.03 \pm 2.70}$  \\
			\multicolumn{3}{l}{{\sc higgs}}                                                                                                                              \\
			$1000\times 28$ & Slisemap           & ${\bf53.31 \pm 2.79}$   & ${\bf0.01 \pm 0.00}$ & $0.02 \pm 0.00$      & $0.99 \pm 0.01$      & $466.24 \pm 175.18$    \\
							& PCA                & $272.96 \pm 1.68$       & $0.21 \pm 0.00$      & $0.26 \pm 0.00$      & $0.47 \pm 0.04$      & ${\bf39.49 \pm 5.00}$  \\
							& Spectral Embedding & $273.49 \pm 3.30$       & $0.20 \pm 0.01$      & $0.26 \pm 0.01$      & $0.50 \pm 0.05$      & ${\bf38.67 \pm 12.46}$ \\
							& LLE                & $274.42 \pm 2.91$       & $0.21 \pm 0.01$      & $0.27 \pm 0.01$      & $0.45 \pm 0.07$      & ${\bf41.70 \pm 5.80}$  \\
							& MLLE               & $282.10 \pm 2.94$       & $0.26 \pm 0.01$      & $0.28 \pm 0.01$      & $0.31 \pm 0.13$      & $44.61 \pm 8.69$       \\
							& MDS                & $272.83 \pm 2.30$       & $0.20 \pm 0.01$      & $0.26 \pm 0.00$      & $0.50 \pm 0.03$      & $111.24 \pm 26.22$     \\
							& Non-Metric MDS     & $277.48 \pm 2.21$       & $0.20 \pm 0.01$      & $0.26 \pm 0.00$      & $0.47 \pm 0.03$      & $47.12 \pm 20.36$      \\
							& Isomap             & $273.36 \pm 3.29$       & $0.21 \pm 0.01$      & $0.26 \pm 0.01$      & $0.49 \pm 0.06$      & $43.38 \pm 12.98$      \\
							& t-SNE              & $274.02 \pm 3.54$       & $0.20 \pm 0.01$      & $0.26 \pm 0.01$      & $0.50 \pm 0.05$      & $47.52 \pm 7.16$       \\
							& UMAP               & $287.73 \pm 2.51$       & $0.27 \pm 0.01$      & $0.28 \pm 0.01$      & $0.30 \pm 0.11$      & ${\bf26.69 \pm 15.38}$ \\
							& Supervised UMAP    & $74.12 \pm 17.22$       & ${\bf0.01 \pm 0.00}$ & ${\bf0.01 \pm 0.00}$ & ${\bf1.00 \pm 0.00}$ & $42.54 \pm 9.91$       \\
			\multicolumn{3}{l}{{\sc higgs (xai)}}                                                                                                                        \\
			$1000\times 28$ & Slisemap           & ${\bf115.33 \pm 10.45}$ & ${\bf0.04 \pm 0.00}$ & ${\bf0.10 \pm 0.01}$ & ${\bf0.86 \pm 0.03}$ & $304.91 \pm 132.32$    \\
							& PCA                & $742.42 \pm 37.94$      & $0.43 \pm 0.02$      & $0.73 \pm 0.04$      & $0.35 \pm 0.01$      & $5.68 \pm 1.82$        \\
							& Spectral Embedding & $744.08 \pm 36.40$      & $0.41 \pm 0.02$      & $0.76 \pm 0.03$      & $0.35 \pm 0.01$      & ${\bf4.81 \pm 1.33}$   \\
							& LLE                & $852.65 \pm 65.06$      & $0.62 \pm 0.08$      & $0.89 \pm 0.07$      & $0.32 \pm 0.02$      & ${\bf5.24 \pm 1.37}$   \\
							& MLLE               & $878.56 \pm 49.85$      & $0.77 \pm 0.07$      & $0.87 \pm 0.05$      & $0.32 \pm 0.02$      & ${\bf4.79 \pm 0.84}$   \\
							& MDS                & $714.88 \pm 33.53$      & $0.37 \pm 0.02$      & $0.72 \pm 0.03$      & $0.36 \pm 0.01$      & $45.55 \pm 9.82$       \\
							& Non-Metric MDS     & $847.28 \pm 49.05$      & $0.48 \pm 0.03$      & $0.88 \pm 0.05$      & $0.32 \pm 0.01$      & ${\bf5.39 \pm 1.60}$   \\
							& Isomap             & $767.62 \pm 41.30$      & $0.46 \pm 0.03$      & $0.79 \pm 0.04$      & $0.34 \pm 0.01$      & $5.96 \pm 1.04$        \\
							& t-SNE              & $781.60 \pm 42.49$      & $0.44 \pm 0.03$      & $0.80 \pm 0.04$      & $0.34 \pm 0.01$      & $6.20 \pm 0.93$        \\
							& UMAP               & $946.18 \pm 54.98$      & $0.87 \pm 0.05$      & $0.90 \pm 0.05$      & $0.31 \pm 0.00$      & $10.28 \pm 2.29$       \\
							& Supervised UMAP    & $947.67 \pm 55.75$      & $0.88 \pm 0.05$      & $0.91 \pm 0.05$      & $0.31 \pm 0.01$      & ${\bf5.14 \pm 1.77}$   \\
			\multicolumn{3}{l}{{\sc covertype}}                                                                                                                          \\
			$1000\times 54$ & Slisemap           & ${\bf54.02 \pm 0.80}$   & ${\bf0.01 \pm 0.00}$ & ${\bf0.01 \pm 0.00}$ & ${\bf1.00 \pm 0.00}$ & $102.80 \pm 24.85$     \\
							& PCA                & $277.74 \pm 1.67$       & $0.25 \pm 0.00$      & $0.26 \pm 0.00$      & $0.55 \pm 0.03$      & ${\bf49.00 \pm 17.60}$ \\
							& Spectral Embedding & $280.70 \pm 3.26$       & $0.25 \pm 0.01$      & $0.27 \pm 0.00$      & $0.47 \pm 0.06$      & ${\bf53.53 \pm 11.50}$ \\
							& LLE                & $280.52 \pm 3.03$       & $0.26 \pm 0.01$      & $0.27 \pm 0.01$      & $0.47 \pm 0.07$      & ${\bf55.00 \pm 11.14}$ \\
							& MLLE               & $283.49 \pm 2.03$       & $0.26 \pm 0.01$      & $0.27 \pm 0.00$      & $0.49 \pm 0.09$      & ${\bf57.72 \pm 20.50}$ \\
							& MDS                & $277.15 \pm 1.76$       & $0.24 \pm 0.00$      & $0.27 \pm 0.00$      & $0.49 \pm 0.03$      & $140.04 \pm 37.21$     \\
							& Non-Metric MDS     & $289.03 \pm 1.16$       & $0.27 \pm 0.01$      & $0.28 \pm 0.00$      & $0.34 \pm 0.04$      & ${\bf47.23 \pm 27.84}$ \\
							& Isomap             & $275.04 \pm 2.42$       & $0.24 \pm 0.00$      & $0.27 \pm 0.00$      & $0.51 \pm 0.04$      & ${\bf58.53 \pm 12.97}$ \\
							& t-SNE              & $271.09 \pm 1.89$       & $0.23 \pm 0.00$      & $0.27 \pm 0.00$      & $0.48 \pm 0.04$      & $61.10 \pm 13.71$      \\
							& UMAP               & $274.85 \pm 2.00$       & $0.24 \pm 0.00$      & $0.27 \pm 0.00$      & $0.48 \pm 0.05$      & ${\bf52.72 \pm 7.74}$  \\
							& Supervised UMAP    & $156.68 \pm 28.05$      & $0.06 \pm 0.02$      & $0.09 \pm 0.03$      & $0.91 \pm 0.04$      & ${\bf53.95 \pm 8.53}$  \\
			\multicolumn{3}{l}{{\sc covertype (xai)}}                                                                                                                    \\
			$1000\times 54$ & Slisemap           & ${\bf70.47 \pm 2.76}$   & ${\bf0.01 \pm 0.00}$ & ${\bf0.03 \pm 0.01}$ & ${\bf0.85 \pm 0.03}$ & $402.65 \pm 143.24$    \\
							& PCA                & $324.37 \pm 21.30$      & $0.26 \pm 0.02$      & $0.28 \pm 0.02$      & $0.32 \pm 0.01$      & ${\bf12.56 \pm 3.34}$  \\
							& Spectral Embedding & $321.61 \pm 22.26$      & $0.25 \pm 0.02$      & $0.30 \pm 0.03$      & $0.32 \pm 0.01$      & ${\bf15.67 \pm 4.47}$  \\
							& LLE                & $336.89 \pm 11.54$      & $0.28 \pm 0.01$      & $0.34 \pm 0.02$      & $0.30 \pm 0.01$      & ${\bf12.40 \pm 4.18}$  \\
							& MLLE               & $327.49 \pm 19.67$      & $0.27 \pm 0.02$      & $0.37 \pm 0.18$      & $0.31 \pm 0.01$      & ${\bf14.94 \pm 5.54}$  \\
							& MDS                & $307.56 \pm 16.67$      & $0.21 \pm 0.01$      & $0.27 \pm 0.02$      & $0.33 \pm 0.01$      & $61.04 \pm 10.45$      \\
							& Non-Metric MDS     & $321.57 \pm 16.29$      & $0.16 \pm 0.01$      & $0.29 \pm 0.02$      & $0.33 \pm 0.01$      & $19.10 \pm 5.41$       \\
							& Isomap             & $318.53 \pm 18.38$      & $0.24 \pm 0.02$      & $0.28 \pm 0.02$      & $0.32 \pm 0.01$      & ${\bf13.26 \pm 3.56}$  \\
							& t-SNE              & $321.57 \pm 16.86$      & $0.23 \pm 0.02$      & $0.29 \pm 0.02$      & $0.32 \pm 0.01$      & $16.91 \pm 4.30$       \\
							& UMAP               & $329.05 \pm 17.01$      & $0.25 \pm 0.02$      & $0.29 \pm 0.02$      & $0.32 \pm 0.01$      & $23.32 \pm 4.25$       \\
							& Supervised UMAP    & $329.91 \pm 19.46$      & $0.26 \pm 0.02$      & $0.29 \pm 0.02$      & $0.32 \pm 0.01$      & $18.09 \pm 4.70$       \\
			\bottomrule
		\end{tabular}
	\end{table}

\end{appendix}

\end{document}